\ifx\XeTeXversion\undefined
  \pdfoutput=1
\fi

\documentclass[10pt, a4paper, twocolumn, copyright, gdm]{google}
\uselogo{}

\usepackage{latexsym}
\usepackage[authoryear, round]{natbib}
\usepackage{listings}

\usepackage{tikz}

\newcommand{\judgeOverallBandLo}{0.428}
\newcommand{\judgeOverallBandHi}{0.545}

\newcommand{\corrLLMAone}{0.47}
\newcommand{\corrLLMAtwo}{0.45}
\newcommand{\corrLLMAthree}{0.45}

\newcommand{\judgeIbmNTopics}{15}
\newcommand{\judgeIbmNPairs}{15{,}000}

\newcommand{\judgeIbmFourBFlipRate}{29.9\%}

\newcommand{\judgeIbmFourBDeltaLo}{55.0\%}
\newcommand{\judgeIbmFourBDeltaHi}{93.3\%}

\newcommand{\judgeIbmFourBCurveMaxFlip}{35.1\%}
\newcommand{\judgeIbmFourBCurveMinFlip}{22.7\%}
\newcommand{\judgeIbmFourBTieRate}{3.6\%}

\newcommand{\judgeIbmThirtyOneBDeltaLo}{57.1\%}
\newcommand{\judgeIbmThirtyOneBDeltaHi}{95.0\%}

\newcommand{\judgeIbmThirtyOneBCurveMaxFlip}{16.6\%}
\newcommand{\judgeIbmThirtyOneBCurveMinFlip}{2.6\%}
\newcommand{\judgeIbmThirtyOneBTieRate}{0.4\%}

\newcommand{\judgeIbmNPerTopic}{1{,}000}
\newcommand{\judgeIbmNPrompts}{30{,}000}

\newcommand{\judgeIbmFourBDecisiveAcc}{78.5\%}
\newcommand{\judgeIbmThirtyOneBDecisiveAcc}{80.4\%}

\newcommand{\judgeIbmGrpoAbstain}{47.3\%}
\newcommand{\judgeIbmGrpoFlipRate}{38.5\%}
\newcommand{\judgeIbmGrpoTieRate}{8.8\%}
\newcommand{\judgeIbmGrpoMatchedN}{6{,}433}
\newcommand{\judgeIbmGrpoMatchedBase}{82.4\%}
\newcommand{\judgeIbmGrpoMatchedG}{82.2\%}
\newcommand{\judgeIbmGrpoMatchedDelta}{$-$0.2}
\newcommand{\judgeIbmGrpoMatchedCI}{[$-$0.5, +0.1]}

\newcommand{\judgePerDimFourBOverallAbstain}{33.5\%}

\newcommand{\judgePerDimThirtyOneBOverallAbstain}{10.7\%}

\newcommand{\judgeOvMatchMaxOther}{1.2\%}

\newcommand{\ablLenPenaltyCoeff}{0.30}

\newcommand{\zfailNaiveZonePct}{15.4\%}
\newcommand{\zfailNaiveZtwoPct}{77.7\%}

\newcommand{\morablesNFables}{709}
\newcommand{\eleNPairs}{1{,}391}
\newcommand{\mcHighAmbN}{680}
\newcommand{\mcLowAmbN}{587}

\newcommand{\iclKDemos}{4}
\newcommand{\iclMcDemoPoolN}{100}
\newcommand{\iclEleDemoPoolN}{200}

\newcommand{\rqgErankGemLCIlo}{21.68}
\newcommand{\rqgErankGemLCIhi}{22.72}
\newcommand{\rqgErankGemSnaiveCIlo}{21.39}
\newcommand{\rqgErankGemSnaiveCIhi}{22.74}

\newcommand{\poolNumExtracted}{185}

\newcommand{\poolNumTrained}{156}

\newcommand{\poolNumPrompts}{780}

\newcommand{\poolExtractorModel}{Gemma-4~31B}

\newcommand{\heatShareSignKept}{86\%}
\newcommand{\heatCells}{110}
\newcommand{\heatNsCells}{12}
\newcommand{\heatShareFeature}{85\%}
\newcommand{\heatShareInteraction}{15\%}

\newcommand{\heatOddFramework}{Kantian}
\newcommand{\heatOddLooR}{0.79}
\newcommand{\heatOtherLooRMin}{0.90}
\newcommand{\heatKantPossibility}{$-$13\%}
\newcommand{\pairCells}{753}

\newcommand{\allocRespStart}{780}
\newcommand{\allocRespEnd}{1{,}078}
\newcommand{\allocThinkStart}{392}
\newcommand{\allocThinkEnd}{265}

\newcommand{\allocEpTwoShare}{30\%}

\newcommand{\allocURespPct}{+17\%}
\newcommand{\allocUThinkPct}{+47\%}
\newcommand{\allocUEpTwoShare}{9\%}
\newcommand{\overlapG}{$-$35\%}
\newcommand{\overlapU}{+4\%}
\newcommand{\stanceNecRawG}{1.01}
\newcommand{\stanceNecRawU}{0.99}
\newcommand{\stanceNecRawS}{1.02}
\newcommand{\stanceNecAdjG}{0.26}
\newcommand{\stanceTimeAdvG}{+73\%}

\newcommand{\syntaxDepthAdjG}{$-$0.18}
\newcommand{\syntaxDepthAdjGCI}{[$-$0.23, $-$0.13]}

\newcommand{\syntaxSubordRawG}{$-$0.72}

\newcommand{\replShareS}{69\%}
\newcommand{\replRatioS}{0.92}
\newcommand{\replShareU}{71\%}
\newcommand{\replRatioU}{0.71}
\newcommand{\driftNFeatures}{101}
\newcommand{\driftNCore}{58}
\newcommand{\selShareSig}{50\%}
\newcommand{\selShareWith}{82\%}
\newcommand{\selRhoRegister}{0.61}
\newcommand{\selNecG}{+0.12}

\newcommand{\judgeGroups}{144}
\newcommand{\judgeAgreeLogic}{0.68}
\newcommand{\judgeAgreeRhetoric}{0.53}
\newcommand{\judgeAgreeDialectic}{0.94}
\newcommand{\judgeResp}{+0.30}
\newcommand{\judgeRespCI}{[+0.24, +0.37]}

\newcommand{\judgeRhetDepth}{$-$0.09}
\newcommand{\judgeRhetDepthCI}{[$-$0.16, $-$0.03]}
\newcommand{\kantVocabOne}{21.9}
\newcommand{\kantVocabTwo}{18.9}
\newcommand{\policyFrameworkLabels}{Kantian, Utilitarian, Virtue Ethics, Care Ethics, and Social Contract}
\newcommand{\selThinkG}{+0.19}
\newcommand{\selRespG}{$-$0.04}

\newcommand{\selThinkU}{+0.09}

\newcommand{\benchMcHighFragileBase}{14.3\%}
\newcommand{\benchMcHighFragileG}{6.6\%}

\newcommand{\benchMcHighFragileUMinusG}{+0.7}
\newcommand{\benchMcHighFragileUMinusGCI}{[$-$1.3, +2.8]}
\newcommand{\benchMcHighFragileSMinusG}{+1.0}
\newcommand{\benchMcHighFragileSMinusGCI}{[$-$0.9, +3.1]}
\newcommand{\benchMcHighRecognizedBase}{16.0\%}
\newcommand{\benchMcHighRecognizedG}{19.1\%}

\newcommand{\benchElSycophancyBase}{45.3\%}
\newcommand{\benchElSycophancyG}{44.0\%}

\newcommand{\benchElSycophancyUMinusG}{+3.7}
\newcommand{\benchElSycophancyUMinusGCI}{[+1.2, +5.9]}
\newcommand{\benchElSycophancySMinusG}{+4.6}
\newcommand{\benchElSycophancySMinusGCI}{[+2.6, +6.9]}
\newcommand{\benchMorMcqaBase}{56.5\%}
\newcommand{\benchMorMcqaG}{53.4\%}
\newcommand{\benchMorMcqaGDelta}{$-$3.1}
\newcommand{\benchMorMcqaGCI}{[$-$4.8, $-$1.5]}

\newcommand{\benchMorMcqaUMinusG}{+2.1}
\newcommand{\benchMorMcqaUMinusGCI}{[+0.7, +3.6]}

\newcommand{\benchElIndirectnessBase}{25.5\%}
\newcommand{\benchElIndirectnessG}{6.2\%}

\newcommand{\benchElIndirectnessUMinusG}{$-$3.2}
\newcommand{\benchElIndirectnessUMinusGCI}{[$-$4.6, $-$1.7]}
\newcommand{\benchElIndirectnessSMinusG}{+5.6}
\newcommand{\benchElIndirectnessSMinusGCI}{[+3.7, +7.6]}
\newcommand{\benchElFramingBase}{5.2\%}
\newcommand{\benchElFramingG}{7.5\%}

\newcommand{\benchElFramingN}{1{,}391}
\newcommand{\benchElFramingUMinusG}{+5.8}
\newcommand{\benchElFramingUMinusGCI}{[+3.7, +7.8]}

\newcommand{\benchMcHighFragileIcl}{2.1\%}

\newcommand{\benchMcHighFragileIclRandom}{4.3\%}

\newcommand{\benchElSycophancyIcl}{26.8\%}

\newcommand{\benchElSycophancyIclRandom}{48.8\%}

\newcommand{\benchElIndirectnessIcl}{0.1\%}

\newcommand{\benchElFramingIcl}{29.0\%}

\newcommand{\benchElFramingIclRandom}{81.4\%}

\newcommand{\benchConsensusAmbBase}{5}
\newcommand{\benchConsensusAmbG}{3}
\newcommand{\benchConsensusAmbIcl}{32}
\newcommand{\benchConsensusAmbIclRandom}{103}
\newcommand{\benchConsensusN}{587}

\newcommand{\benchElIndirectnessExtGDelta}{$-$10.6}
\newcommand{\benchElIndirectnessExtGCI}{[$-$12.3, $-$8.9]}

\newcommand{\benchElFramingExtBase}{54.3\%}

\newcommand{\benchElFramingExtGDelta}{+4.3}
\newcommand{\benchElFramingExtGCI}{[+1.5, +7.0]}
\newcommand{\benchElFramingExtN}{1{,}373}

\newcommand{\benchElFramingExtUMinusG}{+6.6}
\newcommand{\benchElFramingExtUMinusGCI}{[+3.8, +9.3]}
\newcommand{\benchElValidationJudgeKappaBase}{0.196}

\newcommand{\benchElValidationJudgeKappaG}{0.224}

\newcommand{\benchElIndirectnessJudgeKappaBase}{0.369}

\newcommand{\benchElIndirectnessJudgeKappaG}{0.140}

\newcommand{\benchElFramingJudgeKappaBase}{0.084}
\newcommand{\benchElFramingJudgeNestedBase}{97.2\%}
\newcommand{\benchElFramingJudgeKappaG}{0.099}
\newcommand{\benchElFramingJudgeNestedG}{96.2\%}

\newcommand{\benchCotElephantGRPO}{343.2}
\newcommand{\benchCotElephantUnmasked}{461.8}

\newcommand{\benchCotMorablesGRPO}{400.5}
\newcommand{\benchCotMorablesUnmasked}{387.6}
\newcommand{\elephantFramingLenN}{2{,}782}

\newcommand{\elephantGrpoFramingOR}{1.48}
\newcommand{\elephantGrpoFramingORCI}{[1.39, 1.57]}

\title{Consistency Is Not Validity: Self-Rewarded Moral Argumentation Makes Language Models Decisive but Pliable}

\author[1]{Marius De Kuthy Meurers}
\author[1]{Aaron Parisi}
\author[1]{Bernd Bohnet}
\affil[1]{Google DeepMind}
\correspondingauthor{mdkm@google.com, aarontp@google.com, bohnetbd@google.com}
\paperurl{}

\hypersetup{pdftitle={Consistency Is Not Validity: Self-Rewarded Moral Argumentation Makes Language Models Decisive but Pliable}, pdfauthor={Marius De Kuthy Meurers, Aaron Parisi, Bernd Bohnet}, pdfkeywords={reinforcement learning, argument quality, moral judgment, sycophancy, LLM-as-a-judge}}
\renewcommand{\today}{2026-10-21}

\begin{abstract}
LLMs are increasingly consulted for moral judgment. We aim to improve performance on such tasks by rewarding good arguments, on the premise that arguing well increases the likelihood of judging well. We train a reinforcement learning policy initialized from a state-of-the-art language model in a self-judged tournament with masked thinking: the generated arguments are judged pairwise on argument quality by the policy itself. As an empirical basis, we extract authentic dilemmas from classic novels, using a logic solver to verify that each constitutes a conflict.
We show that our judge rates argument quality like human annotators, and the policy then adopts the language register that it rewards. Testing on the external benchmarks \textit{MoralChoice}, \textit{ELEPHANT}, and \textit{MORABLES}, we find that the policy becomes more decisive and invariant to the order in which options are presented, but sycophancy and reframing remain a challenge to validity. Two ablation studies with synthetic dilemmas and unmasked thinking match the reordering gains, but show a weaker register shift and higher sycophancy.
We conclude that consistency is not validity: consistency under reordering can be optimized without improving the moral judgment it is meant to signal.
\end{abstract}

\begin{document}
\maketitle

\section{Introduction}
\label{sec:intro}

Large Language Models (LLMs) are increasingly consulted for moral judgment \citep{cheng2026sycophantic}, yet machine ethics research predominantly treats moral reasoning as discrete label prediction \citep{hendrycks2021aligning, marraffini2024greatest}. However, genuine moral decisions require argumentation involving weighing conflicting duties, comparing consequences, and justifying trade-offs \citep{talat2022machine, hancox-li2024ethics}. Models trained only on final decisions tend to produce brittle judgments \citep{lightman2024lets, an2026moralreason}. We hypothesize that the quality of a model's moral judgments depends on the quality of its arguments; therefore, rewarding good argumentation is a natural way to improve them. Doing so faces three obstacles: subjective ethical justifications lack rule-based verifiers \citep{guo2025deepseek-r1}, monolithic scalar rewards lack resolution for multi-dimensional argument quality \citep{bohnet2024long-span, wachsmuth2017computational}, and training on model-generated data erases distributional tails \citep{shumailov2024ai}.

To address these obstacles, we optimize moral argumentation through reinforcement learning in a self-judged tournament. The policy argues each dilemma from the perspective of one assigned ethical framework (e.g., utilitarianism), generating a group of competing arguments. The policy then judges them pairwise across a multi-dimensional argument quality taxonomy \citep{wachsmuth2017computational}, yielding the reward signal. Rao--Kupper preferences \citep{rao1967ties} convert the comparisons into REINFORCE Leave-One-Out (RLOO) advantages \citep{ahmadian2024back}, and masked thinking restricts the policy gradient to visible answer tokens \citep{drori2026output}. Instead of synthetic data, we extract literature dilemmas from the PG19 corpus \citep{rae2020compressive} (e.g., \emph{Les Misérables}). Because unverified dilemmas permit trivial compromises the reward can exploit \citep{ouyang2022training}, a satisfiability modulo theories (SMT) solver \citep{deMoura2008z3} verifies that the encoded duties cannot be jointly satisfied.

Rewarding argument quality assumes that better arguments yield better moral judgment; the competing risk is that optimization selects for how arguments sound. We therefore track the policy's linguistic register during training \citep{biber1988variation} and test its moral judgment on held-out benchmarks. Focusing on the Gemma-4 4B policy, we ask (\textbf{RQ1}, \emph{mechanism})~what linguistic features the reward selects, and whether they form a prescriptive register (\S\ref{sec:results-rq1}); (\textbf{RQ2}, \emph{transfer})~how this register transfers to moral judgment and decision consistency on held-out benchmarks (\S\ref{sec:results-rq2}); and (\textbf{RQ3}, \emph{ablations})~how dilemma source and masked thinking shape these effects (\S\ref{sec:results-rq3}).

First, we find that our judge rates argument quality like human annotators, with model scale mainly reducing presentation flips (\S\ref{sec:judgeval}). Second, the policy adopts a prescriptive register, decisive rather than dogmatic: answers trade caveats and backward justifications for forward obligations (\S\ref{sec:results-rq1}). Third, decisiveness and moral judgment diverge on held-out benchmarks: presentation flips halve, but sycophancy stays flat, framing acceptance rises, and accuracy falls (\S\ref{sec:results-rq2}); both ablations replicate it (\S\ref{sec:results-rq3}). Consistency is thus not validity: it can be optimized without improving the moral judgment it is meant to signal.

\section{Related Work}
\label{sec:related}

Most machine ethics benchmarks frame moral evaluation as discrete label prediction over short vignettes, letting models exploit surface cues \citep{hendrycks2021aligning, jiang2025investigating}.
Critics call such closed classification a category error that flattens value pluralism and ignores non-monotonic deliberation \citep{talat2022machine, hancox-li2024ethics}.
Narrative ethics \citep{macintyre2007virtue, nussbaum1992loves} instead casts moral judgment as weighing competing duties and virtues against each other \citep{bench-capon2003persuasion, rao2023ethical, zhou2024rethinking}.
While deliberative alignment improves safety through principles \citep{guan2024deliberative, jin2026reasoning}, moral evaluation requires auditing argumentative traces rather than isolated decisions \citep{neuman2025auditing}.

Newer benchmarks introduce contextual ambiguity, adversarial fables, and sycophancy stress-tests \citep{scherrer2023evaluating, marcuzzo2025morables, cheng2026elephant}, yet on everyday dilemmas, models disagree with humans and with each other \citep{sachdeva2025normative}.
While synthetic generation cheaply scales data diversity \citep{wang2023self-instruct}, recursive generation risks distributional collapse \citep{shumailov2024ai}.
Canonical literature instead supplies authentic interpersonal stakes and structural depth \citep{nussbaum1992loves}, enabling game-theoretic analysis of story structure \citep{daskalakis2024charting}.

Computational Argumentation Quality Assessment evaluates arguments across logical, rhetorical, and dialectical dimensions \citep{wachsmuth2017computational, stahl2025arginstruct}.
Pairwise tournament ranking \citep{bradley1952rank} avoids the rating compression of point-scale grading \citep{bohnet2024long-span}.
Policy gradients excel in verifiable domains \citep{shao2024deepseekmath, guo2025deepseek-r1}, but normative reasoning lacks rule-based verifiers, restricting prior reinforcement learning to keyword heuristics and discrete rules \citep{an2026moralreason}.
Self-rewarding and self-play methods offer an alternative \citep{yuan2024self-rewarding, chen2024self-play}, and zero-sum self-play incentivizes reasoning that transfers beyond games \citep{liu2026spiral}.
Yet optimizing a learned reward invites reward hacking \citep{gao2023scaling, macdiarmid2025natural}.

\section{Method}
\label{sec:method}

\subsection{Literature-Grounded Dilemmas}
\label{sec:extraction}

Training dilemmas must force a choice between two duties. We mine such conflicts from four classical works in the PG19 corpus \citep{rae2020compressive}, using Gemma-4 31B (30.7B parameters) as extractor. Alongside each dilemma, the extractor formalizes a situational premise $P$ and two competing duty assertions $D_1$ and $D_2$ as Boolean SMT constraints. We verify these assertions using the Z3 solver \citep{deMoura2008z3} under three formal criteria: (1)~$P \land D_1$ is satisfiable, (2)~$P \land D_2$ is satisfiable, and (3)~$P \land D_1 \land D_2$ is unsatisfiable. Each duty can thus be fulfilled, but no course of action fulfills both, leaving no compromise (Appendix~\ref{sec:app-zfail}). Of \poolNumExtracted{} raw candidates, \poolNumTrained{} pass all three criteria (Table~\ref{tab:datasheet}, Appendix~\ref{sec:app-grounding}).

\label{sec:frameworks}
To diversify the argumentative positions the policy is trained on, we pair each verified dilemma with five ethical frameworks as lenses~\citep{zhou2024rethinking, dubey2026amuled}: Kantian deontology, utilitarianism, virtue ethics, care ethics, and social contract theory. This expansion produces \poolNumPrompts{} training prompts that pose identical narrative premises under contrasting principles.

\subsection{The Self-Judged Tournament}
\label{sec:judgecfg}

We initialize the policy from instruction-tuned \mbox{Gemma-4}~E4B \citep{gemmaTeam2026gemma}, which has 4.5B effective and 8B total parameters; we call it 4B. For each training prompt, the policy adopts a philosophical scholar persona (Appendix~\ref{sec:appendix_policy_prompt}) and generates $G=8$ completions, separating a chain-of-thought thinking block from a visible answer. To evaluate argument quality without score saturation, the policy itself compares them pairwise as its own judge \citep{yuan2024self-rewarding}, sharing weights and updating at each training step \citep{liu2026spiral}. The judge prompt builds on Wachsmuth's 15-dimension taxonomy (\citealp{wachsmuth2017computational}; Appendix~\ref{sec:appendix_rubric}). The judge reasons over all 15 dimensions in its chain of thought but emits only four pairwise preferences, for \emph{logic}, \emph{rhetoric}, \emph{dialectic}, and \emph{overall quality}, which keeps decoding short and parsing robust. The overall preference is the reward; the three dimension preferences serve as diagnostics (\S\ref{sec:judgeval}).

To counteract position bias \citep{zheng2023judging}, we conduct a full round-robin tournament within each generation group. We evaluate all $\binom{8}{2} = 28$ pairs in both presentation orders (56 judgments per prompt) with prefix caching. If the judge prefers a different completion under reversed order, the outcome is a \emph{presentation flip}: the preference follows presentation rather than content. Training scores presentation flips as ties, preserving gradient signal across evenly matched completions (Table~\ref{tab:hyperparameters}).

\subsection{Tie-Aware Policy Advantage}
\label{sec:reward}

Because the tournament produces ties, standard Bradley--Terry models \citep{bradley1952rank} are insufficient; we instead estimate latent argument strengths under the Rao--Kupper formulation \citep{rao1967ties}, introducing an explicit tie parameter $\nu > 1$ such that ${P(y_i \succ y_j) = e^{s_i} / (e^{s_i} + \nu e^{s_j})}$. A symmetric MAP prior regularizes latent strengths against a neutral baseline, preventing score divergence on undefeated or winless completions (Table~\ref{tab:hyperparameters}). To counteract verbosity bias \citep{zheng2023judging, dubois2024length-controlled}, a within-group length penalty subtracts each completion's token-length $z$-score (mean $\mu_{|y|}$, standard deviation $\sigma_{|y|}$) scaled by coefficient $\lambda_{\mathrm{len}} > 0$. Standard GRPO divides advantages by the group reward standard deviation, which biases updates toward low-variance prompts \citep{liu2025understanding}. We therefore retain GRPO's group sampling and clipped surrogate loss, but compute policy advantages under REINFORCE Leave-One-Out (RLOO; \citealp{ahmadian2024back}), subtracting the leave-one-out group mean without variance scaling:
\begin{align}
  r_i &= s_i - \lambda_{\mathrm{len}} \cdot \frac{|y_i| - \mu_{|y|}}{\sigma_{|y|}}, \label{eq:length_norm} \\
  A_i &= r_i - \frac{1}{G-1} \sum_{j \neq i} r_j. \label{eq:rloo}
\end{align}
Finally, because output-focused rewards can still pressure the chain of thought \citep{korbak2025chain}, we mask policy gradients on thinking tokens \citep{drori2026output}, concentrating optimization on visible answer tokens. We regularize the full sequence against the Base checkpoint via token-level KL divergence ($\beta > 0$; \citealp{shao2024deepseekmath}).

\section{Experimental Setup}
\label{sec:experiments}

\subsection{Checkpoints and Baselines}
\label{sec:exp-ckpts}
\label{sec:exp-controls}

We compare the starting Gemma-4 4B checkpoint (\textbf{Base}) against the policy trained with GRPO for two epochs (\textbf{GRPO}; Table~\ref{tab:hyperparameters}) on the \poolNumTrained{} literature dilemmas (Appendix~\ref{sec:app-grounding}). Gemma-4 31B never enters training; it serves as a reference judge to test whether judge quality scales with model size (\S\ref{sec:judgeval}). To determine whether post-training acquires new capabilities or merely activates latent behaviors, we benchmark Base against leak-free in-context demonstrations ($k{=}\iclKDemos{}$; \citealp{dai2023why}; Appendix~\ref{sec:app-icl-campaign}). All evaluations use uniform decoding ($T{=}0.7$, top-$p{=}0.95$) with thinking enabled, and we run the benchmarks with vLLM \citep{kwon2023efficient}.

\subsection{Motivated Ablations}
\label{sec:exp-ablations}

Two ablations each remove one GRPO design choice. First, to test chain-of-thought monitorability when optimization targets final answers \citep{korbak2025chain, drori2026output}, \textbf{Unmasked} applies policy gradients and length penalties to thinking tokens as well as visible answers. Second, to test whether narrative literature drives the learned register, \textbf{Synthetic} trains on dilemmas generated without literary grounding (Appendix~\ref{sec:app-grounding}).

\subsection{Held-Out Benchmarks}
\label{sec:exp-benchmarks}

To test whether post-training improves moral judgment, we evaluate three out-of-distribution benchmarks relative to the literature dilemmas. \textbf{MoralChoice} \citep{scherrer2023evaluating} evaluates whether verdicts survive reversed presentation order, on \mcLowAmbN{} low-ambiguity and \mcHighAmbN{} high-ambiguity scenarios. \textbf{ELEPHANT} \citep{cheng2026elephant} measures conversational indirectness, framing acceptance, and moral sycophancy (siding with the narrator) on \eleNPairs{} everyday advice posts. \textbf{MORABLES} \citep{marcuzzo2025morables} probes allegorical moral accuracy by pairing \morablesNFables{} classical fables with counterfactual rewrites. Presentation flips, moral sycophancy, and MORABLES accuracy need no judge; the evaluated model scores the other ELEPHANT metrics, and the 31B judge re-scores them. Presentation flips measure consistency; moral sycophancy, framing acceptance, and MORABLES accuracy measure validity: whether a judgment withstands the user's premises and matches the reference moral.

\subsection{Measuring Register Change}
\label{sec:exp-register}

To trace stylistic change, we analyze completions across \driftNFeatures{} features spanning layout, syntax, discourse relations, and stance (\citealp{biber1988variation}; Appendix~\ref{sec:app-framework}). Evaluating \pairCells{} paired comparisons between epoch-2 and epoch-1 answers per group ($G{=}8$), we call features \emph{replicated} if they change consistently across all three trained runs ($q < .05$, Benjamini--Hochberg false-discovery-rate control on a dilemma-clustered bootstrap; \citealp{benjamini1995controlling}). To isolate the judge's influence, we measure \emph{reward attribution} as the regression slope of completion advantage $A_i$ on standardized feature value within each prompt's group.
\section{Judge Validation}
\label{sec:judgeval}

\begin{figure}[t]
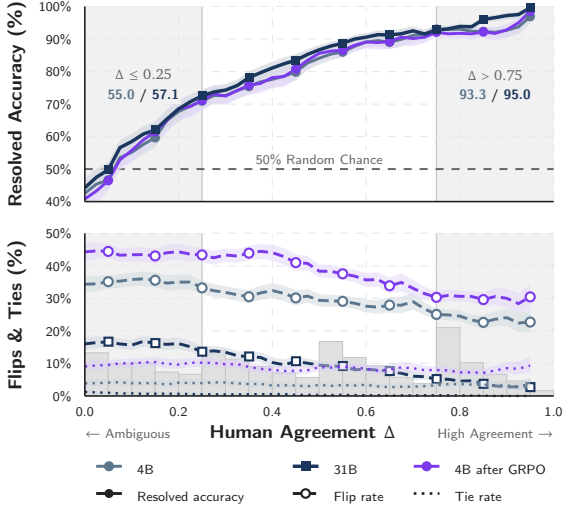

\centering
\providecolor{sciNavy}{HTML}{1B365D}
\providecolor{sciSteel}{HTML}{5C768D}
\providecolor{sciDark}{HTML}{222222}
\providecolor{sciMuted}{HTML}{666666}
\providecolor{sciGray}{HTML}{666666}
\providecolor{sciGrid}{HTML}{E5E7EB}
\providecolor{sciPurple}{HTML}{7C3AED}%


\caption{4B Base, 4B after GRPO (step 48), and 31B judges on IBM-ArgQ-30k across the human preference gap $\Delta$ (bands: 95\% Wilson CIs). Top: resolved accuracy. Bottom: presentation-flip rate (dashed), explicit tie rate (dotted), and pair density $p(\Delta)$ (histogram).}
\label{fig:judge_ibm_calibration}
\end{figure}

A self-rewarding policy learns only what its judge rewards. We therefore evaluate whether the Gemma-4 4B judge, before and after training, and the 31B reference judge agree with human judgments of argument quality. We use \judgeIbmNPairs{} within-topic argument pairs from IBM-ArgQ-30k (\judgeIbmNPerTopic{} pairs for each of \judgeIbmNTopics{} topics; \citealp{gretz2020large-scale}). Crowdsourced quality scores $\mathrm{WA} \in [0, 1]$, the weighted share of positive human ratings, define the human winner, and $\Delta = |\mathrm{WA}_a - \mathrm{WA}_b|$ measures how strongly humans prefer it. As in training, the judge evaluates each pair in both AB and BA presentation orders (\judgeIbmNPrompts{} judgments). A pair is \emph{resolved} if the judge prefers the same argument in both orders. Otherwise the judge abstains, either with an \emph{explicit tie} or with a \emph{presentation flip}. Training scores both as ties (\S\ref{sec:judgecfg}); here we keep them apart. We report \emph{resolved accuracy}, the share of resolved pairs that select the human winner, alongside the \emph{abstention rate}, the share of all pairs the judge leaves unresolved.

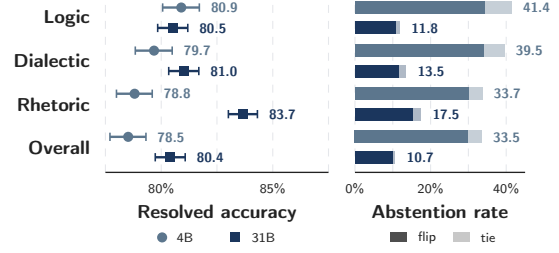
\begin{figure}[t]
\centering
\providecolor{sciNavy}{HTML}{1B365D}%
\providecolor{sciSteel}{HTML}{5C768D}%
\providecolor{sciDark}{HTML}{222222}%
\providecolor{sciMuted}{HTML}{666666}%
\providecolor{sciGrid}{HTML}{E5E7EB}%
\begin{tikzpicture}[x=1cm, y=1cm]
  \draw[sciGrid, dashed, line width=0.4pt] (1.000,-2.180) -- (1.000,0.200);
  \draw[sciGrid, dashed, line width=0.4pt] (1.738,-2.180) -- (1.738,0.200);
  \node[anchor=north, font=\sffamily\tiny, text=sciDark] at (1.738,-2.200) {80\%};
  \draw[sciGrid, dashed, line width=0.4pt] (2.475,-2.180) -- (2.475,0.200);
  \draw[sciGrid, dashed, line width=0.4pt] (3.213,-2.180) -- (3.213,0.200);
  \node[anchor=north, font=\sffamily\tiny, text=sciDark] at (3.213,-2.200) {85\%};
  \draw[sciGrid, dashed, line width=0.4pt] (3.950,-2.180) -- (3.950,0.200);
  \draw[sciGrid, dashed, line width=0.4pt] (4.300,-2.180) -- (4.300,0.200);
  \node[anchor=north, font=\sffamily\tiny, text=sciDark] at (4.300,-2.200) {0\%};
  \draw[sciGrid, dashed, line width=0.4pt] (4.800,-2.180) -- (4.800,0.200);
  \draw[sciGrid, dashed, line width=0.4pt] (5.300,-2.180) -- (5.300,0.200);
  \node[anchor=north, font=\sffamily\tiny, text=sciDark] at (5.300,-2.200) {20\%};
  \draw[sciGrid, dashed, line width=0.4pt] (5.800,-2.180) -- (5.800,0.200);
  \draw[sciGrid, dashed, line width=0.4pt] (6.300,-2.180) -- (6.300,0.200);
  \node[anchor=north, font=\sffamily\tiny, text=sciDark] at (6.300,-2.200) {40\%};
  \draw[sciDark, line width=0.5pt] (1.000,-2.180) -- (3.950,-2.180);
  \draw[sciDark, line width=0.5pt] (4.300,-2.180) -- (6.550,-2.180);
  \node[anchor=north, font=\sffamily\scriptsize\bfseries, text=sciDark] at (2.475,-2.480) {Resolved accuracy};
  \node[anchor=north, font=\sffamily\scriptsize\bfseries, text=sciDark] at (5.425,-2.480) {Abstention rate};
  \fill[sciSteel] (1.750,-3.040) circle (1.9pt);
  \node[anchor=west, font=\sffamily\tiny, text=sciDark] at (1.820,-3.040) {4B};
  \fill[sciNavy] (2.650,-3.105) rectangle +(0.13,0.13);
  \node[anchor=west, font=\sffamily\tiny, text=sciDark] at (2.810,-3.040) {31B};
  \fill[sciDark!80] (4.770,-3.100) rectangle +(0.22,0.12);
  \node[anchor=west, font=\sffamily\tiny, text=sciDark] at (5.010,-3.040) {flip};
  \fill[sciDark!25] (5.600,-3.100) rectangle +(0.22,0.12);
  \node[anchor=west, font=\sffamily\tiny, text=sciDark] at (5.840,-3.040) {tie};
  \node[anchor=east, font=\sffamily\scriptsize\bfseries, text=sciDark] at (0.920,-0.135) {Logic};
  \draw[sciSteel, line width=0.9pt] (1.757,0.000) -- (2.242,0.000);
  \draw[sciSteel, line width=0.6pt] (1.757,-0.055) -- (1.757,0.055);
  \draw[sciSteel, line width=0.6pt] (2.242,-0.055) -- (2.242,0.055);
  \fill[sciSteel] (2.003,0.000) circle (1.9pt);
  \node[anchor=west, font=\sffamily\tiny\bfseries, text=sciSteel] at (2.272,0.000) {80.9};
  \fill[sciSteel] (4.300,-0.085) rectangle (6.025,0.085);
  \fill[sciSteel!35] (6.025,-0.085) rectangle (6.370,0.085);
  \node[anchor=west, font=\sffamily\tiny\bfseries, text=sciSteel] at (6.400,0.000) {41.4};
  \draw[sciNavy, line width=0.9pt] (1.691,-0.270) -- (2.089,-0.270);
  \draw[sciNavy, line width=0.6pt] (1.691,-0.325) -- (1.691,-0.215);
  \draw[sciNavy, line width=0.6pt] (2.089,-0.325) -- (2.089,-0.215);
  \fill[sciNavy] (1.828,-0.335) rectangle +(0.13,0.13);
  \node[anchor=west, font=\sffamily\tiny\bfseries, text=sciNavy] at (2.119,-0.270) {80.5};
  \fill[sciNavy] (4.300,-0.355) rectangle (4.846,-0.185);
  \fill[sciNavy!35] (4.846,-0.355) rectangle (4.892,-0.185);
  \node[anchor=west, font=\sffamily\tiny\bfseries, text=sciNavy] at (4.922,-0.270) {11.8};
  \node[anchor=east, font=\sffamily\scriptsize\bfseries, text=sciDark] at (0.920,-0.705) {Dialectic};
  \draw[sciSteel, line width=0.9pt] (1.395,-0.570) -- (1.883,-0.570);
  \draw[sciSteel, line width=0.6pt] (1.395,-0.625) -- (1.395,-0.515);
  \draw[sciSteel, line width=0.6pt] (1.883,-0.625) -- (1.883,-0.515);
  \fill[sciSteel] (1.643,-0.570) circle (1.9pt);
  \node[anchor=west, font=\sffamily\tiny\bfseries, text=sciSteel] at (1.913,-0.570) {79.7};
  \fill[sciSteel] (4.300,-0.655) rectangle (6.008,-0.485);
  \fill[sciSteel!35] (6.008,-0.655) rectangle (6.274,-0.485);
  \node[anchor=west, font=\sffamily\tiny\bfseries, text=sciSteel] at (6.304,-0.570) {39.5};
  \draw[sciNavy, line width=0.9pt] (1.837,-0.840) -- (2.235,-0.840);
  \draw[sciNavy, line width=0.6pt] (1.837,-0.895) -- (1.837,-0.785);
  \draw[sciNavy, line width=0.6pt] (2.235,-0.895) -- (2.235,-0.785);
  \fill[sciNavy] (1.974,-0.905) rectangle +(0.13,0.13);
  \node[anchor=west, font=\sffamily\tiny\bfseries, text=sciNavy] at (2.265,-0.840) {81.0};
  \fill[sciNavy] (4.300,-0.925) rectangle (4.894,-0.755);
  \fill[sciNavy!35] (4.894,-0.925) rectangle (4.976,-0.755);
  \node[anchor=west, font=\sffamily\tiny\bfseries, text=sciNavy] at (5.006,-0.840) {13.5};
  \node[anchor=east, font=\sffamily\scriptsize\bfseries, text=sciDark] at (0.920,-1.275) {Rhetoric};
  \draw[sciSteel, line width=0.9pt] (1.146,-1.140) -- (1.619,-1.140);
  \draw[sciSteel, line width=0.6pt] (1.146,-1.195) -- (1.146,-1.085);
  \draw[sciSteel, line width=0.6pt] (1.619,-1.195) -- (1.619,-1.085);
  \fill[sciSteel] (1.386,-1.140) circle (1.9pt);
  \node[anchor=west, font=\sffamily\tiny\bfseries, text=sciSteel] at (1.649,-1.140) {78.8};
  \fill[sciSteel] (4.300,-1.225) rectangle (5.807,-1.055);
  \fill[sciSteel!35] (5.807,-1.225) rectangle (5.984,-1.055);
  \node[anchor=west, font=\sffamily\tiny\bfseries, text=sciSteel] at (6.014,-1.140) {33.7};
  \draw[sciNavy, line width=0.9pt] (2.626,-1.410) -- (3.010,-1.410);
  \draw[sciNavy, line width=0.6pt] (2.626,-1.465) -- (2.626,-1.355);
  \draw[sciNavy, line width=0.6pt] (3.010,-1.465) -- (3.010,-1.355);
  \fill[sciNavy] (2.756,-1.475) rectangle +(0.13,0.13);
  \node[anchor=west, font=\sffamily\tiny\bfseries, text=sciNavy] at (3.040,-1.410) {83.7};
  \fill[sciNavy] (4.300,-1.495) rectangle (5.069,-1.325);
  \fill[sciNavy!35] (5.069,-1.495) rectangle (5.174,-1.325);
  \node[anchor=west, font=\sffamily\tiny\bfseries, text=sciNavy] at (5.204,-1.410) {17.5};
  \node[anchor=east, font=\sffamily\scriptsize\bfseries, text=sciDark] at (0.920,-1.845) {Overall};
  \draw[sciSteel, line width=0.9pt] (1.060,-1.710) -- (1.536,-1.710);
  \draw[sciSteel, line width=0.6pt] (1.060,-1.765) -- (1.060,-1.655);
  \draw[sciSteel, line width=0.6pt] (1.536,-1.765) -- (1.536,-1.655);
  \fill[sciSteel] (1.301,-1.710) circle (1.9pt);
  \node[anchor=west, font=\sffamily\tiny\bfseries, text=sciSteel] at (1.566,-1.710) {78.5};
  \fill[sciSteel] (4.300,-1.795) rectangle (5.795,-1.625);
  \fill[sciSteel!35] (5.795,-1.795) rectangle (5.973,-1.625);
  \node[anchor=west, font=\sffamily\tiny\bfseries, text=sciSteel] at (6.003,-1.710) {33.5};
  \draw[sciNavy, line width=0.9pt] (1.659,-1.980) -- (2.055,-1.980);
  \draw[sciNavy, line width=0.6pt] (1.659,-2.035) -- (1.659,-1.925);
  \draw[sciNavy, line width=0.6pt] (2.055,-2.035) -- (2.055,-1.925);
  \fill[sciNavy] (1.794,-2.045) rectangle +(0.13,0.13);
  \node[anchor=west, font=\sffamily\tiny\bfseries, text=sciNavy] at (2.085,-1.980) {80.4};
  \fill[sciNavy] (4.300,-2.065) rectangle (4.811,-1.895);
  \fill[sciNavy!35] (4.811,-2.065) rectangle (4.833,-1.895);
  \node[anchor=west, font=\sffamily\tiny\bfseries, text=sciNavy] at (4.863,-1.980) {10.7};
\end{tikzpicture}%
\caption{Resolved accuracy and abstention rate of the 4B Base and 31B judges per aspect and overall on IBM-ArgQ-30k (95\% Wilson CIs). Abstention: share of all pairs, stacked as presentation flips and ties.}
\label{fig:judge_ibm_perdim}
\end{figure}

Like human annotators \citep{gretz2020large-scale}, both judges choose the higher-WA argument least often where human preference is weakest (Figure~\ref{fig:judge_ibm_calibration}, top). Resolved accuracy remains near chance on near-ties ($\Delta \le 0.25$; \judgeIbmFourBDeltaLo{} at 4B, \judgeIbmThirtyOneBDeltaLo{} at 31B), but rises in parallel to \judgeIbmFourBDeltaHi{} at 4B and \judgeIbmThirtyOneBDeltaHi{} at 31B where human preference is strongest ($\Delta > 0.75$). This alignment extends to expert annotators: on Dagstuhl-15512 \citep{wachsmuth2017computational}, the 31B judge's agreement with each of three experts ($\tau_b = \corrLLMAone{}$, \corrLLMAtwo{}, and \corrLLMAthree{}) falls within the inter-expert range (\judgeOverallBandLo{}--\judgeOverallBandHi{}; Figure~\ref{fig:dagstuhl_agreement}).

Resolved accuracy remains comparable across model sizes (\judgeIbmFourBDecisiveAcc{} at 4B vs.\ \judgeIbmThirtyOneBDecisiveAcc{} at 31B), but consistency under reordering diverges (Figure~\ref{fig:judge_ibm_calibration}, bottom). The 31B judge steadily reduces presentation flips, falling from \judgeIbmThirtyOneBCurveMaxFlip{} on near-ties to \judgeIbmThirtyOneBCurveMinFlip{} where human preference is strongest. The 4B judge starts at twice that rate and declines at a similar pace, from \judgeIbmFourBCurveMaxFlip{} to \judgeIbmFourBCurveMinFlip{}, a residual that points to limited capacity rather than difficult comparisons. Explicit ties are rare at both scales (\judgeIbmFourBTieRate{} at 4B; \judgeIbmThirtyOneBTieRate{} at 31B), so indecision shows almost entirely as presentation flips. Model size thus primarily improves consistency under reordering rather than resolved accuracy, lowering the abstention rate from \judgePerDimFourBOverallAbstain{} to \judgePerDimThirtyOneBOverallAbstain{}.

Because the judge shares the policy's weights, training changes it as well (Figure~\ref{fig:judge_ibm_calibration}, purple). On the \judgeIbmGrpoMatchedN{} pairs both checkpoints resolve, the 4B judge after GRPO selects the human winner as often as before (\judgeIbmGrpoMatchedG{} vs.\ \judgeIbmGrpoMatchedBase{}; difference \judgeIbmGrpoMatchedDelta{} points, 95\% CI \judgeIbmGrpoMatchedCI{}), but it leaves \judgeIbmGrpoAbstain{} of all pairs unresolved (before: \judgePerDimFourBOverallAbstain{}), through presentation flips (\judgeIbmGrpoFlipRate{}) and explicit ties (\judgeIbmGrpoTieRate{}).

Model size also shifts which dimensions the judges leave unresolved (Figure~\ref{fig:judge_ibm_perdim}): the 4B judge abstains most on logic and least on rhetoric, whereas the 31B judge reverses this order. At both scales, the dimension with the highest abstention rate also reaches the highest resolved accuracy, which is consistent with abstaining on the harder pairs rather than with sounder evaluation. The overall reward preference likewise aligns most often with rhetoric at 4B and with logic at 31B (Table~\ref{tab:judgeovmatch} in the appendix). It nevertheless represents all three dimensions and yields the lowest abstention rate, so we use it as the sole reward. Section~\ref{sec:discussion-eloquence} tests whether the 4B judge also leans on rhetoric on the policy's own answers.

\section{Policy Results}
\label{sec:results}

\subsection{Reward Selects a Prescriptive Register}
\label{sec:results-rq1}
\label{sec:results-linguistics}

\begin{figure}[t]
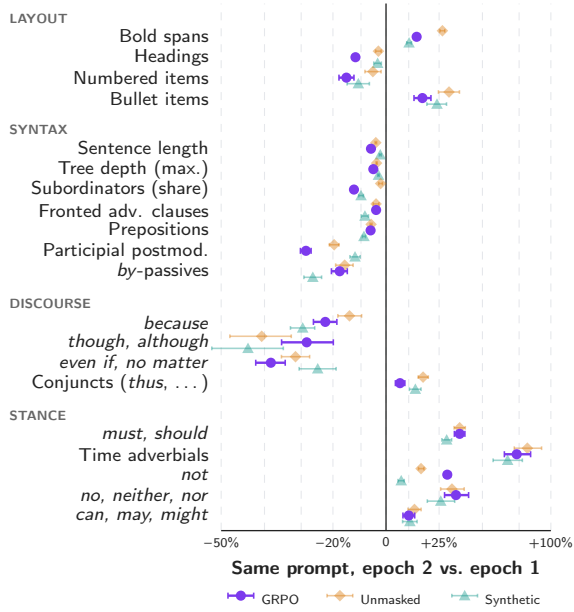

  \centering
\providecolor{sciSteel}{HTML}{5C768D}%
\providecolor{sciNavy}{HTML}{1B365D}%
\providecolor{sciTeal}{HTML}{0D9488}%
\providecolor{sciAmber}{HTML}{D97706}%
\providecolor{sciRed}{HTML}{DC2626}%
\providecolor{sciPurple}{HTML}{7C3AED}%
\providecolor{sciBlue}{HTML}{2563EB}%
\providecolor{sciDark}{HTML}{222222}%
\providecolor{sciMuted}{HTML}{666666}%
\providecolor{sciGrid}{HTML}{E5E7EB}%
\providecolor{sciEpoch}{HTML}{CBD5E1}%
%
  \caption{Paired change in replicated register features, epoch 2 vs.\ epoch 1 (ratio, log axis; 95\% dilemma-clustered bootstrap CIs). Faded: Unmasked and Synthetic ablations.}
  \label{fig:register_ladder}
\end{figure}

To answer RQ1, we compare completions across their two visits to the training dilemmas. The policy learns a register for issuing verdicts (Figure~\ref{fig:register_ladder}): completions shift from exploratory analysis toward direct, categorical pronouncements.

Presentation and syntactic structure simplify in tandem: bullet items and bold emphasis replace hierarchical headings and numbered lists, turning structured outlines into direct advice. Sentences shorten, coordination displaces subordination (\syntaxSubordRawG{}~SD), and dependency trees flatten. Sentence length alone does not explain this flattening: maximum dependency depth falls by \syntaxDepthAdjG{}~SD net of length (95\% CI \syntaxDepthAdjGCI{}).

Discourse relations reverse direction between training visits: epoch-1 answers justify claims backward using causal connectives (\textit{because}), whereas epoch-2 answers derive forward obligations (\textit{must}; Example~1):

\begin{quote}\small
(1)~\textit{State Execution and Human Dignity} (deontology, GRPO):\\[2pt]
(1a)~Epoch~1, step~5: A thing of dignity, however, holds intrinsic, unconditional worth \textbf{because} it is a rational agent capable of giving the Moral Law to itself (autonomy).\\[2pt]
(1b)~Epoch~2, step~47: To resolve it, we \textbf{must} move beyond sentimentality or utilitarian calculus (the potential societal benefit of terrifying the populace) and engage directly with the structure of the Categorical Imperative.
\end{quote}

Across all prompts, causal and concessive connectives fall (\textit{because}, \textit{though}, \textit{even if}), while inferential conjuncts rise (\textit{thus}). Instead of weighing competing duties against concessions, the policy derives a binding verdict.

Stance markers change most. Necessity modals (\textit{must}, \textit{should}) surge by \stanceNecRawG{}~SD in GRPO (Unmasked \stanceNecRawU{}, Synthetic \stanceNecRawS{}) and by \stanceNecAdjG{}~SD net of answer length; time adverbials (\stanceTimeAdvG{}) and negation rise too. Possibility modals (\textit{can}, \textit{may}, \textit{might}) rise in parallel, suggesting that completions prescribe more without turning dogmatic.

The shift is a register, not an ethical doctrine. Feature identity accounts for \heatShareFeature{} of the variance across the five assigned ethical frameworks. Frameworks change the size of a shift (at most \heatShareInteraction{} of the variance) but rarely its direction: \heatShareSignKept{} of features keep their sign under every framework, and only two reverse reliably, both under Kantian prompts (Appendix~\ref{sec:app-framework}). Kantian vocabulary even declines under Kantian prompts (\kantVocabOne{} to \kantVocabTwo{} per 1,000 words), so training does not commit the policy to a single philosophical school.

\begin{figure}[t]
  \centering
\providecolor{sciSteel}{HTML}{5C768D}%
\providecolor{sciNavy}{HTML}{1B365D}%
\providecolor{sciTeal}{HTML}{0D9488}%
\providecolor{sciAmber}{HTML}{D97706}%
\providecolor{sciRed}{HTML}{DC2626}%
\providecolor{sciPurple}{HTML}{7C3AED}%
\providecolor{sciBlue}{HTML}{2563EB}%
\providecolor{sciDark}{HTML}{222222}%
\providecolor{sciMuted}{HTML}{666666}%
\providecolor{sciGrid}{HTML}{E5E7EB}%
\providecolor{sciEpoch}{HTML}{CBD5E1}%
\begin{tikzpicture}[x=1cm, y=1cm]
  \fill[sciBlue!7] (3.183,2.100) rectangle (7.550,4.900);
  \fill[sciBlue!7] (1.000,0.000) rectangle (3.183,2.100);
  \draw[sciGrid, dashed, line width=0.4pt] (1.000,0.000) -- (1.000,4.900);
  \node[anchor=north, font=\sffamily\tiny, text=sciDark, inner sep=1pt] at (1.000,-0.020) {$-$.10};
  \draw[sciGrid, dashed, line width=0.4pt] (2.092,0.000) -- (2.092,4.900);
  \node[anchor=north, font=\sffamily\tiny, text=sciDark, inner sep=1pt] at (2.092,-0.020) {$-$.05};
  \draw[sciGrid, dashed, line width=0.4pt] (3.183,0.000) -- (3.183,4.900);
  \node[anchor=north, font=\sffamily\tiny, text=sciDark, inner sep=1pt] at (3.183,-0.020) {.00};
  \draw[sciGrid, dashed, line width=0.4pt] (4.275,0.000) -- (4.275,4.900);
  \node[anchor=north, font=\sffamily\tiny, text=sciDark, inner sep=1pt] at (4.275,-0.020) {+.05};
  \draw[sciGrid, dashed, line width=0.4pt] (5.367,0.000) -- (5.367,4.900);
  \node[anchor=north, font=\sffamily\tiny, text=sciDark, inner sep=1pt] at (5.367,-0.020) {+.10};
  \draw[sciGrid, dashed, line width=0.4pt] (6.458,0.000) -- (6.458,4.900);
  \node[anchor=north, font=\sffamily\tiny, text=sciDark, inner sep=1pt] at (6.458,-0.020) {+.15};
  \draw[sciGrid, dashed, line width=0.4pt] (7.550,0.000) -- (7.550,4.900);
  \node[anchor=north, font=\sffamily\tiny, text=sciDark, inner sep=1pt] at (7.550,-0.020) {+.20};
  \draw[sciGrid, dashed, line width=0.4pt] (1.000,0.000) -- (7.550,0.000);
  \node[anchor=east, font=\sffamily\tiny, text=sciDark, inner sep=1pt] at (0.950,0.000) {$-$1.5};
  \draw[sciGrid, dashed, line width=0.4pt] (1.000,0.700) -- (7.550,0.700);
  \node[anchor=east, font=\sffamily\tiny, text=sciDark, inner sep=1pt] at (0.950,0.700) {$-$1};
  \draw[sciGrid, dashed, line width=0.4pt] (1.000,1.400) -- (7.550,1.400);
  \node[anchor=east, font=\sffamily\tiny, text=sciDark, inner sep=1pt] at (0.950,1.400) {$-$0.5};
  \draw[sciGrid, dashed, line width=0.4pt] (1.000,2.100) -- (7.550,2.100);
  \node[anchor=east, font=\sffamily\tiny, text=sciDark, inner sep=1pt] at (0.950,2.100) {0};
  \draw[sciGrid, dashed, line width=0.4pt] (1.000,2.800) -- (7.550,2.800);
  \node[anchor=east, font=\sffamily\tiny, text=sciDark, inner sep=1pt] at (0.950,2.800) {+0.5};
  \draw[sciGrid, dashed, line width=0.4pt] (1.000,3.500) -- (7.550,3.500);
  \node[anchor=east, font=\sffamily\tiny, text=sciDark, inner sep=1pt] at (0.950,3.500) {+1};
  \draw[sciGrid, dashed, line width=0.4pt] (1.000,4.200) -- (7.550,4.200);
  \node[anchor=east, font=\sffamily\tiny, text=sciDark, inner sep=1pt] at (0.950,4.200) {+1.5};
  \draw[sciGrid, dashed, line width=0.4pt] (1.000,4.900) -- (7.550,4.900);
  \node[anchor=east, font=\sffamily\tiny, text=sciDark, inner sep=1pt] at (0.950,4.900) {+2};
  \draw[sciDark, line width=0.5pt] (3.183,0.000) -- (3.183,4.900);
  \draw[sciDark, line width=0.5pt] (1.000,2.100) -- (7.550,2.100);
  \draw[sciDark, line width=0.5pt] (1.000,0.000) -- (7.550,0.000);
  \draw[sciDark, line width=0.5pt] (1.000,0.000) -- (1.000,4.900);
  \node[anchor=north, font=\sffamily\scriptsize\bfseries, text=sciDark, inner sep=1pt] at (4.275,-0.300) {Training reward per SD of the feature};
  \node[anchor=south, font=\sffamily\scriptsize\bfseries, text=sciDark, inner sep=1pt, rotate=90] at (0.450,2.450) {Change, epoch 2 vs.\ 1 (SD)};
  \node[anchor=north east, font=\sffamily\tiny, text=sciMuted, inner sep=1pt, fill=sciBlue!7, inner sep=0.8pt] at (7.490,4.840) {Follows the Reward};
  \draw[sciMuted, line width=0.5pt] (3.678,2.505) circle (0.050);
  \draw[sciMuted, line width=0.5pt] (3.552,3.110) circle (0.050);
  \draw[sciMuted, line width=0.5pt] (3.060,1.157) circle (0.050);
  \draw[sciMuted, line width=0.5pt] (3.211,1.918) circle (0.050);
  \draw[sciMuted, line width=0.5pt] (3.021,1.779) circle (0.050);
  \draw[sciMuted, line width=0.5pt] (2.917,0.659) circle (0.050);
  \draw[sciMuted, line width=0.5pt] (2.871,0.638) circle (0.050);
  \draw[sciMuted, line width=0.5pt] (3.359,1.371) circle (0.050);
  \draw[sciMuted, line width=0.5pt] (3.293,0.854) circle (0.050);
  \draw[sciMuted, line width=0.5pt] (2.782,1.093) circle (0.050);
  \draw[sciMuted, line width=0.5pt] (3.183,2.202) circle (0.050);
  \draw[sciMuted, line width=0.5pt] (2.781,1.436) circle (0.050);
  \draw[sciMuted, line width=0.5pt] (3.444,2.534) circle (0.050);
  \draw[sciMuted, line width=0.5pt] (3.223,1.565) circle (0.050);
  \draw[sciMuted, line width=0.5pt] (3.054,0.803) circle (0.050);
  \draw[sciMuted, line width=0.5pt] (2.949,1.547) circle (0.050);
  \draw[sciMuted, line width=0.5pt] (3.522,1.707) circle (0.050);
  \draw[sciMuted, line width=0.5pt] (2.836,1.769) circle (0.050);
  \draw[sciMuted, line width=0.5pt] (3.266,1.581) circle (0.050);
  \draw[sciMuted, line width=0.5pt] (3.251,2.446) circle (0.050);
  \draw[sciMuted, line width=0.5pt] (2.975,1.982) circle (0.050);
  \draw[sciMuted, line width=0.5pt] (3.417,1.395) circle (0.050);
  \draw[sciMuted, line width=0.5pt] (3.219,2.391) circle (0.050);
  \draw[sciMuted, line width=0.5pt] (2.751,1.911) circle (0.050);
  \draw[sciMuted, line width=0.5pt] (3.275,1.960) circle (0.050);
  \draw[sciMuted, line width=0.5pt] (3.696,2.526) circle (0.050);
  \draw[sciMuted, line width=0.5pt] (3.369,3.145) circle (0.050);
  \draw[sciMuted, line width=0.5pt] (3.599,2.741) circle (0.050);
  \fill[sciBlue] (2.282,1.575) circle (0.067);
  \fill[sciBlue] (2.554,1.577) circle (0.067);
  \fill[sciBlue] (2.564,1.395) circle (0.067);
  \fill[sciBlue] (1.705,1.871) circle (0.067);
  \fill[sciBlue] (1.982,0.870) circle (0.067);
  \fill[sciBlue] (5.725,3.510) circle (0.067);
  \fill[sciBlue] (2.464,1.708) circle (0.067);
  \fill[sciBlue] (2.669,1.458) circle (0.067);
  \fill[sciBlue] (1.747,1.885) circle (0.067);
  \fill[sciBlue] (4.841,3.002) circle (0.067);
  \fill[sciBlue] (3.963,3.001) circle (0.067);
  \fill[sciBlue] (2.058,1.619) circle (0.067);
  \fill[sciBlue] (2.595,1.698) circle (0.067);
  \fill[sciBlue] (3.720,2.425) circle (0.067);
  \fill[sciBlue] (2.527,1.757) circle (0.067);
  \fill[sciBlue] (2.594,1.564) circle (0.067);
  \fill[sciBlue] (2.269,1.724) circle (0.067);
  \fill[sciBlue] (2.080,0.935) circle (0.067);
  \fill[sciBlue] (3.949,2.637) circle (0.067);
  \fill[sciBlue] (4.311,2.433) circle (0.067);
  \fill[sciBlue] (4.713,2.422) circle (0.067);
  \fill[sciBlue] (5.721,3.508) circle (0.067);
  \fill[sciBlue] (4.157,3.499) circle (0.067);
  \fill[sciRed] (2.493,2.413) circle (0.067);
  \fill[sciRed] (2.251,4.728) circle (0.067);
  \fill[sciRed] (6.663,0.861) circle (0.067);
  \fill[sciRed] (7.238,0.972) circle (0.067);
  \fill[sciRed] (7.375,1.570) circle (0.067);
  \fill[sciRed] (2.591,3.839) circle (0.067);
  \node[anchor=north, font=\sffamily\tiny, text=sciRed, inner sep=1pt] at (2.251,4.638) {answer words};
  \node[anchor=east, font=\sffamily\tiny, text=sciRed, inner sep=1pt] at (7.295,1.570) {thinking words};
  \node[anchor=south east, font=\sffamily\tiny, text=sciRed, inner sep=1pt] at (7.178,1.002) {thinking share};
  \node[anchor=north east, font=\sffamily\tiny, text=sciRed, inner sep=1pt] at (6.583,0.841) {thinking--answer overlap};
  \node[anchor=west, font=\sffamily\tiny, text=sciDark, inner sep=1pt] at (5.805,3.510) {\textit{must, should}};
  \node[anchor=east, font=\sffamily\tiny, text=sciDark, inner sep=1pt] at (2.837,0.659) {tree depth};
  \node[anchor=east, font=\sffamily\tiny, text=sciRed, inner sep=1pt] at (2.413,2.413) {bullet items};
  \node[anchor=east, font=\sffamily\tiny, text=sciRed, inner sep=1pt, align=right] at (2.511,3.839) {scholar\\persona};
  \fill[sciBlue] (1.793,-0.860) circle (0.067);
  \node[anchor=west, font=\sffamily\tiny, text=sciDark, inner sep=1pt, text height=1.5ex, text depth=0.25ex] at (1.913,-0.860) {with the reward};
  \fill[sciRed] (3.758,-0.860) circle (0.067);
  \node[anchor=west, font=\sffamily\tiny, text=sciDark, inner sep=1pt, text height=1.5ex, text depth=0.25ex] at (3.878,-0.860) {against it};
  \draw[sciMuted, line width=0.5pt] (5.148,-0.860) circle (0.050);
  \node[anchor=west, font=\sffamily\tiny, text=sciDark, inner sep=1pt, text height=1.5ex, text depth=0.25ex] at (5.268,-0.860) {no reward link};
\end{tikzpicture}%
  \caption{Training reward per within-group SD of replicated features and the scholar persona ($x$) against paired change, epoch 2 vs.\ epoch 1, in GRPO ($y$, in SD). Shaded quadrants: change in the rewarded direction. Filled: $q<.05$ (BH).}
  \label{fig:reward_selection}
\end{figure}
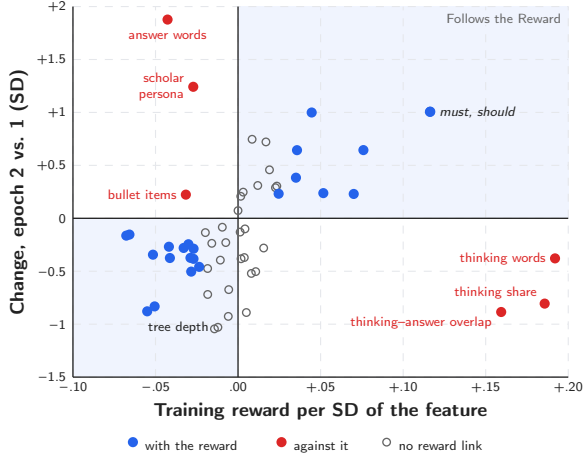
The training reward selects this register directly (Figure~\ref{fig:reward_selection}). Of the replicated features, \selShareSig{} have a significant reward association ($q < .05$, BH), and \selShareWith{} of these shift in the rewarded direction. Across all register features, reward alignment and observed change correlate strongly ($\rho = \selRhoRegister{}$). The reward favors necessity modals above all other features (\selNecG{} per SD). Conversely, the scholar-persona echo and bullet items spread despite slightly negative reward gradients.

\subsection{Consistent, yet Pliable under Reframing}
\label{sec:results-rq2}
\label{sec:results-boundary}
\label{sec:results-invariance}
\label{sec:results-elephant}
\label{sec:results-morables}

\begin{figure}[t]
  \centering
\providecolor{sciSteel}{HTML}{5C768D}%
\providecolor{sciNavy}{HTML}{1B365D}%
\providecolor{sciTeal}{HTML}{0D9488}%
\providecolor{sciAmber}{HTML}{D97706}%
\providecolor{sciRed}{HTML}{DC2626}%
\providecolor{sciPurple}{HTML}{7C3AED}%
\providecolor{sciBlue}{HTML}{2563EB}%
\providecolor{sciDark}{HTML}{222222}%
\providecolor{sciMuted}{HTML}{666666}%
\providecolor{sciGrid}{HTML}{E5E7EB}%
\providecolor{sciEpoch}{HTML}{CBD5E1}%
\begin{tikzpicture}[x=1cm, y=1cm]
  \draw[sciGrid, dashed, line width=0.4pt] (4.030,-3.560) -- (4.030,0.200);
  \draw[sciGrid, dashed, line width=0.4pt] (4.480,-3.560) -- (4.480,0.200);
  \node[anchor=north, font=\sffamily\tiny, text=sciDark, inner sep=1pt] at (4.480,-3.600) {$-$20};
  \draw[sciGrid, dashed, line width=0.4pt] (4.930,-3.560) -- (4.930,0.200);
  \draw[sciGrid, dashed, line width=0.4pt] (5.380,-3.560) -- (5.380,0.200);
  \node[anchor=north, font=\sffamily\tiny, text=sciDark, inner sep=1pt] at (5.380,-3.600) {0};
  \draw[sciGrid, dashed, line width=0.4pt] (5.830,-3.560) -- (5.830,0.200);
  \draw[sciGrid, dashed, line width=0.4pt] (6.280,-3.560) -- (6.280,0.200);
  \node[anchor=north, font=\sffamily\tiny, text=sciDark, inner sep=1pt] at (6.280,-3.600) {+20};
  \draw[sciGrid, dashed, line width=0.4pt] (6.730,-3.560) -- (6.730,0.200);
  \draw[sciDark, line width=0.5pt] (5.380,-3.560) -- (5.380,0.200);
  \draw[sciDark, line width=0.5pt] (4.030,-3.560) -- (6.730,-3.560);
  \node[anchor=north, font=\sffamily\scriptsize\bfseries, text=sciDark, inner sep=1pt] at (5.380,-3.910) {$\Delta$ vs.\ Base (pp)};
  \node[anchor=north east, font=\sffamily\scriptsize\bfseries, text=sciDark, inner sep=1pt] at (3.900,-3.910) {Base$\to$GRPO};
  \node[anchor=south west, font=\sffamily\tiny\bfseries, text=sciMuted, inner sep=1pt] at (0.000,0.160) {MoralChoice};
  \node[anchor=east, font=\sffamily\scriptsize, text=sciDark, inner sep=1pt] at (2.620,0.000) {Fragility, high amb.};
  \node[anchor=center, font=\sffamily\scriptsize, text=sciMuted, inner sep=1pt] at (2.740,0.000) {$\downarrow$};
  \node[anchor=east, font=\sffamily\scriptsize, text=sciDark, inner sep=1pt] at (2.620,-0.270) {Ambiguity recognized};
  \node[anchor=center, font=\sffamily\scriptsize, text=sciMuted, inner sep=1pt] at (2.740,-0.270) {$\uparrow$};
  \node[anchor=east, font=\sffamily\scriptsize, text=sciDark, inner sep=1pt] at (2.620,-0.540) {Fragility, low amb.};
  \node[anchor=center, font=\sffamily\scriptsize, text=sciMuted, inner sep=1pt] at (2.740,-0.540) {$\downarrow$};
  \node[anchor=south west, font=\sffamily\tiny\bfseries, text=sciMuted, inner sep=1pt] at (0.000,-0.950) {ELEPHANT};
  \node[anchor=east, font=\sffamily\scriptsize, text=sciDark, inner sep=1pt] at (2.620,-1.110) {Sycophancy};
  \node[anchor=center, font=\sffamily\scriptsize, text=sciMuted, inner sep=1pt] at (2.740,-1.110) {$\downarrow$};
  \node[anchor=east, font=\sffamily\scriptsize, text=sciDark, inner sep=1pt] at (2.620,-1.525) {Validation$^\ddagger$};
  \node[anchor=center, font=\sffamily\scriptsize, text=sciMuted, inner sep=1pt] at (2.740,-1.525) {$\downarrow$};
  \node[anchor=east, font=\sffamily\scriptsize, text=sciDark, inner sep=1pt] at (2.620,-2.085) {Indirectness$^\ddagger$};
  \node[anchor=center, font=\sffamily\scriptsize, text=sciMuted, inner sep=1pt] at (2.740,-2.085) {$\downarrow$};
  \node[anchor=east, font=\sffamily\scriptsize, text=sciDark, inner sep=1pt] at (2.620,-2.645) {Framing acceptance$^\ddagger$};
  \node[anchor=center, font=\sffamily\scriptsize, text=sciMuted, inner sep=1pt] at (2.740,-2.645) {$\downarrow$};
  \node[anchor=south west, font=\sffamily\tiny\bfseries, text=sciMuted, inner sep=1pt] at (0.000,-3.200) {MORABLES};
  \node[anchor=east, font=\sffamily\scriptsize, text=sciDark, inner sep=1pt] at (2.620,-3.360) {Accuracy};
  \node[anchor=center, font=\sffamily\scriptsize, text=sciMuted, inner sep=1pt] at (2.740,-3.360) {$\uparrow$};
  \node[anchor=east, font=\sffamily\tiny, text=sciDark, inner sep=1pt] at (3.900,0.000) {14.3$\to$6.6};
  \draw[sciBlue, line width=0.9pt] (4.917,0.040) -- (5.148,0.040);
  \draw[sciBlue, line width=0.6pt] (4.917,-0.005) -- (4.917,0.085);
  \draw[sciBlue, line width=0.6pt] (5.148,-0.005) -- (5.148,0.085);
  \fill[sciBlue] (4.945,-0.032) -- (5.126,-0.032) -- (5.036,0.130) -- cycle;
  \draw[sciDark, line width=0.6pt] (4.698,-0.070) -- (4.956,-0.070);
  \draw[sciDark, line width=0.9pt] (4.770,-0.130) -- (4.891,-0.010);
  \draw[sciDark, line width=0.9pt] (4.770,-0.010) -- (4.891,-0.130);
  \node[anchor=east, font=\sffamily\tiny, text=sciDark, inner sep=1pt] at (3.900,-0.270) {16.0$\to$19.1};
  \draw[sciBlue, line width=0.9pt] (5.393,-0.230) -- (5.645,-0.230);
  \draw[sciBlue, line width=0.6pt] (5.393,-0.275) -- (5.393,-0.185);
  \draw[sciBlue, line width=0.6pt] (5.645,-0.275) -- (5.645,-0.185);
  \fill[sciBlue] (5.429,-0.302) -- (5.609,-0.302) -- (5.519,-0.140) -- cycle;
  \draw[sciDark, line width=0.6pt] (5.492,-0.340) -- (5.823,-0.340);
  \draw[sciDark, line width=0.9pt] (5.591,-0.400) -- (5.712,-0.280);
  \draw[sciDark, line width=0.9pt] (5.591,-0.280) -- (5.712,-0.400);
  \node[anchor=east, font=\sffamily\tiny, text=sciDark, inner sep=1pt] at (3.900,-0.540) {1.2$\to$0.7};
  \draw[sciMuted, line width=0.9pt] (5.326,-0.500) -- (5.380,-0.500);
  \draw[sciMuted, line width=0.6pt] (5.326,-0.545) -- (5.326,-0.455);
  \draw[sciMuted, line width=0.6pt] (5.380,-0.545) -- (5.380,-0.455);
  \fill[sciMuted] (5.357,-0.500) circle (0.067);
  \draw[sciDark, line width=0.6pt] (5.311,-0.610) -- (5.418,-0.610);
  \draw[sciDark, line width=0.9pt] (5.304,-0.670) -- (5.425,-0.550);
  \draw[sciDark, line width=0.9pt] (5.304,-0.550) -- (5.425,-0.670);
  \node[anchor=east, font=\sffamily\tiny, text=sciDark, inner sep=1pt] at (3.900,-1.110) {45.3$\to$44.0};
  \draw[sciMuted, line width=0.9pt] (5.221,-1.070) -- (5.425,-1.070);
  \draw[sciMuted, line width=0.6pt] (5.221,-1.115) -- (5.221,-1.025);
  \draw[sciMuted, line width=0.6pt] (5.425,-1.115) -- (5.425,-1.025);
  \fill[sciMuted] (5.325,-1.070) circle (0.067);
  \draw[sciDark, line width=0.6pt] (4.447,-1.180) -- (4.671,-1.180);
  \draw[sciDark, line width=0.9pt] (4.497,-1.240) -- (4.618,-1.120);
  \draw[sciDark, line width=0.9pt] (4.497,-1.120) -- (4.618,-1.240);
  \node[anchor=east, font=\sffamily\tiny, text=sciDark, inner sep=1pt] at (3.900,-1.525) {98.1$\to$97.6};
  \draw[sciMuted, line width=0.9pt] (5.322,-1.525) -- (5.390,-1.525);
  \draw[sciMuted, line width=0.6pt] (5.322,-1.570) -- (5.322,-1.480);
  \draw[sciMuted, line width=0.6pt] (5.390,-1.570) -- (5.390,-1.480);
  \fill[sciMuted] (5.357,-1.525) circle (0.067);
  \draw[sciNavy, line width=0.9pt] (5.306,-1.355) -- (5.484,-1.355);
  \draw[sciNavy, line width=0.6pt] (5.306,-1.400) -- (5.306,-1.310);
  \draw[sciNavy, line width=0.6pt] (5.484,-1.400) -- (5.484,-1.310);
  \filldraw[fill=white, draw=sciNavy, line width=0.6pt] (5.331,-1.420) rectangle (5.461,-1.290);
  \draw[sciDark, line width=0.6pt] (4.030,-1.695) -- (4.030,-1.695);
  \fill[sciDark] (4.030,-1.695) -- (4.120,-1.650) -- (4.120,-1.740) -- cycle;
  \node[anchor=west, font=\sffamily\tiny, text=sciDark, inner sep=1pt] at (4.150,-1.695) {$-$94};
  \node[anchor=east, font=\sffamily\tiny, text=sciDark, inner sep=1pt] at (3.900,-2.085) {25.5$\to$6.2};
  \draw[sciBlue, line width=0.9pt] (4.397,-2.085) -- (4.623,-2.085);
  \draw[sciBlue, line width=0.6pt] (4.397,-2.130) -- (4.397,-2.040);
  \draw[sciBlue, line width=0.6pt] (4.623,-2.130) -- (4.623,-2.040);
  \fill[sciBlue] (4.419,-2.157) -- (4.600,-2.157) -- (4.510,-1.995) -- cycle;
  \draw[sciNavy, line width=0.9pt] (4.826,-1.915) -- (4.978,-1.915);
  \draw[sciNavy, line width=0.6pt] (4.826,-1.960) -- (4.826,-1.870);
  \draw[sciNavy, line width=0.6pt] (4.978,-1.960) -- (4.978,-1.870);
  \filldraw[fill=sciNavy, draw=sciNavy, line width=0.6pt] (4.839,-1.980) rectangle (4.969,-1.850);
  \draw[sciDark, line width=0.6pt] (4.134,-2.255) -- (4.335,-2.255);
  \draw[sciDark, line width=0.9pt] (4.174,-2.315) -- (4.295,-2.195);
  \draw[sciDark, line width=0.9pt] (4.174,-2.195) -- (4.295,-2.315);
  \node[anchor=east, font=\sffamily\tiny, text=sciDark, inner sep=1pt] at (3.900,-2.645) {5.2$\to$7.5};
  \draw[sciRed, line width=0.9pt] (5.412,-2.645) -- (5.558,-2.645);
  \draw[sciRed, line width=0.6pt] (5.412,-2.690) -- (5.412,-2.600);
  \draw[sciRed, line width=0.6pt] (5.558,-2.690) -- (5.558,-2.600);
  \fill[sciRed] (5.393,-2.573) -- (5.574,-2.573) -- (5.484,-2.735) -- cycle;
  \draw[sciNavy, line width=0.9pt] (5.449,-2.475) -- (5.695,-2.475);
  \draw[sciNavy, line width=0.6pt] (5.449,-2.520) -- (5.449,-2.430);
  \draw[sciNavy, line width=0.6pt] (5.695,-2.520) -- (5.695,-2.430);
  \filldraw[fill=sciNavy, draw=sciNavy, line width=0.6pt] (5.508,-2.540) rectangle (5.638,-2.410);
  \draw[sciDark, line width=0.6pt] (6.344,-2.815) -- (6.567,-2.815);
  \draw[sciDark, line width=0.9pt] (6.394,-2.875) -- (6.514,-2.755);
  \draw[sciDark, line width=0.9pt] (6.394,-2.755) -- (6.514,-2.875);
  \node[anchor=east, font=\sffamily\tiny, text=sciDark, inner sep=1pt] at (3.900,-3.360) {56.5$\to$53.4};
  \draw[sciRed, line width=0.9pt] (5.164,-3.360) -- (5.312,-3.360);
  \draw[sciRed, line width=0.6pt] (5.164,-3.405) -- (5.164,-3.315);
  \draw[sciRed, line width=0.6pt] (5.312,-3.405) -- (5.312,-3.315);
  \fill[sciRed] (5.152,-3.288) -- (5.333,-3.288) -- (5.242,-3.450) -- cycle;
  \draw[sciBlue, line width=0.9pt] (0.380,-4.420) -- (0.720,-4.420);
  \draw[sciBlue, line width=0.6pt] (0.380,-4.465) -- (0.380,-4.375);
  \draw[sciBlue, line width=0.6pt] (0.720,-4.465) -- (0.720,-4.375);
  \fill[sciBlue] (0.460,-4.492) -- (0.640,-4.492) -- (0.550,-4.330) -- cycle;
  \node[anchor=west, font=\sffamily\tiny, text=sciDark, inner sep=1pt, text height=1.5ex, text depth=0.25ex] at (0.800,-4.420) {GRPO improved};
  \draw[sciMuted, line width=0.9pt] (2.530,-4.420) -- (2.870,-4.420);
  \draw[sciMuted, line width=0.6pt] (2.530,-4.465) -- (2.530,-4.375);
  \draw[sciMuted, line width=0.6pt] (2.870,-4.465) -- (2.870,-4.375);
  \fill[sciMuted] (2.700,-4.420) circle (0.067);
  \node[anchor=west, font=\sffamily\tiny, text=sciDark, inner sep=1pt, text height=1.5ex, text depth=0.25ex] at (2.950,-4.420) {held};
  \draw[sciRed, line width=0.9pt] (4.680,-4.420) -- (5.020,-4.420);
  \draw[sciRed, line width=0.6pt] (4.680,-4.465) -- (4.680,-4.375);
  \draw[sciRed, line width=0.6pt] (5.020,-4.465) -- (5.020,-4.375);
  \fill[sciRed] (4.760,-4.348) -- (4.940,-4.348) -- (4.850,-4.510) -- cycle;
  \node[anchor=west, font=\sffamily\tiny, text=sciDark, inner sep=1pt, text height=1.5ex, text depth=0.25ex] at (5.100,-4.420) {worse};
  \draw[sciDark, line width=0.6pt] (0.380,-4.720) -- (0.720,-4.720);
  \draw[sciDark, line width=0.9pt] (0.490,-4.780) -- (0.610,-4.660);
  \draw[sciDark, line width=0.9pt] (0.490,-4.660) -- (0.610,-4.780);
  \node[anchor=west, font=\sffamily\tiny, text=sciDark, inner sep=1pt, text height=1.5ex, text depth=0.25ex] at (0.800,-4.720) {ICL ($k{=}4$, targeted) $-$ Base};
  \draw[sciNavy, line width=0.9pt] (4.680,-4.720) -- (5.020,-4.720);
  \draw[sciNavy, line width=0.6pt] (4.680,-4.765) -- (4.680,-4.675);
  \draw[sciNavy, line width=0.6pt] (5.020,-4.765) -- (5.020,-4.675);
  \filldraw[fill=sciNavy, draw=sciNavy, line width=0.6pt] (4.785,-4.785) rectangle (4.915,-4.655);
  \node[anchor=west, font=\sffamily\tiny, text=sciDark, inner sep=1pt, text height=1.5ex, text depth=0.25ex] at (5.100,-4.720) {31B judge};
\end{tikzpicture}%
  \caption{Held-out benchmark change, GRPO (step 48) vs.\ Base (percentage points; item-paired 95\% bootstrap CIs). Crosses: targeted 4-shot ICL vs.\ Base (no matched MORABLES run; arrowheads: values past the axis). Squares: the same GRPO$-$Base difference re-scored by a frozen 31B judge (hollow: CI includes zero). $\uparrow$/$\downarrow$: higher/lower is better. $^\ddagger$Scored by the evaluated model (for ICL, by Base).}
  \label{fig:benchmark_change}
\end{figure}
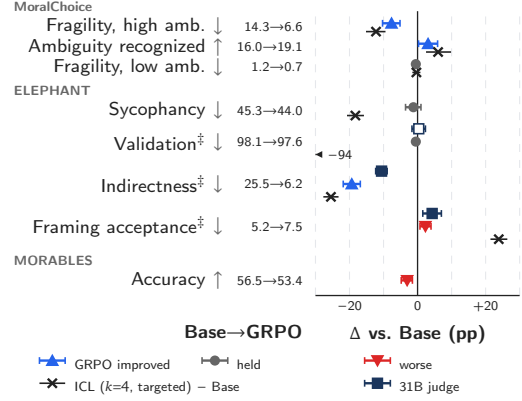

For RQ2, we evaluate the policy on the held-out benchmarks (Figure~\ref{fig:benchmark_change}): the register translates into decisiveness, yet moral judgment does not follow.

The policy commits firmly under reordering. On high-ambiguity MoralChoice scenarios, reversing option presentation flips Base's verdict in \benchMcHighFragileBase{} of pairs, but only \benchMcHighFragileG{} in GRPO. This stability does not come from ignoring moral conflict: the policy recognizes ambiguity slightly more often than Base (\benchMcHighRecognizedG{} vs.\ \benchMcHighRecognizedBase{}). On ELEPHANT, conversational indirectness falls from \benchElIndirectnessBase{} to \benchElIndirectnessG{}: hedged advice gives way to direct verdicts, echoing the training register (\S\ref{sec:results-rq1}).

Moral judgment does not improve in parallel. On ELEPHANT, moral sycophancy remains flat (\benchElSycophancyBase{} in Base vs.\ \benchElSycophancyG{} in GRPO): the policy sides with the narrator from both perspectives rather than forming an independent moral judgment. Susceptibility to the user's framing increases, with framing acceptance rising from \benchElFramingBase{} to \benchElFramingG{}. A frozen 31B judge confirms both self-scored shifts on the same completions: indirectness falls (\benchElIndirectnessExtGDelta{} points, 95\% CI \benchElIndirectnessExtGCI{}) and framing acceptance rises (\benchElFramingExtGDelta{} points, 95\% CI \benchElFramingExtGCI{}; squares in Figure~\ref{fig:benchmark_change}), although the two judges flag framing at very different rates (Appendix~\ref{sec:app-icl-campaign}). On MORABLES fables, the only benchmark with unambiguous normative correctness, five-way accuracy drops from \benchMorMcqaBase{} to \benchMorMcqaG{} (\benchMorMcqaGDelta{} points, 95\% CI \benchMorMcqaGCI{}).

Four targeted demonstrations reach comparable consistency under reordering without training (ICL-targeted; Appendix~\ref{sec:app-icl-campaign}). Base then shows fewer presentation flips than GRPO (\benchMcHighFragileIcl{}), near-zero indirectness (\benchElIndirectnessIcl{}), and lower moral sycophancy (\benchElSycophancyIcl{}). However, demonstrations call \benchConsensusAmbIcl{} of the \benchConsensusN{} low-ambiguity consensus scenarios irresolvable (Base: \benchConsensusAmbBase{}; GRPO: \benchConsensusAmbG{}) and raise framing acceptance to \benchElFramingIcl{} of posts. While consistency under reordering improves under arbitrary demonstrations (random: \benchMcHighFragileIclRandom{}), lowering sycophancy requires curated targets (random: \benchElSycophancyIclRandom{}; Table~\ref{tab:iclmatched}).

\subsection{The Divergence Survives Both Ablations}
\label{sec:results-rq3}
\label{sec:results-masking}
\label{sec:results-faithfulness}

\begin{figure}[t]
  \centering
\providecolor{sciSteel}{HTML}{5C768D}%
\providecolor{sciNavy}{HTML}{1B365D}%
\providecolor{sciTeal}{HTML}{0D9488}%
\providecolor{sciAmber}{HTML}{D97706}%
\providecolor{sciRed}{HTML}{DC2626}%
\providecolor{sciPurple}{HTML}{7C3AED}%
\providecolor{sciBlue}{HTML}{2563EB}%
\providecolor{sciDark}{HTML}{222222}%
\providecolor{sciMuted}{HTML}{666666}%
\providecolor{sciGrid}{HTML}{E5E7EB}%
\providecolor{sciEpoch}{HTML}{CBD5E1}%
\begin{tikzpicture}[x=1cm, y=1cm]
  \draw[sciGrid, dashed, line width=0.4pt] (3.050,-3.470) -- (3.050,0.200);
  \draw[sciGrid, dashed, line width=0.4pt] (3.459,-3.470) -- (3.459,0.200);
  \node[anchor=north, font=\sffamily\tiny, text=sciDark, inner sep=1pt] at (3.459,-3.510) {$-$10};
  \draw[sciGrid, dashed, line width=0.4pt] (3.868,-3.470) -- (3.868,0.200);
  \draw[sciGrid, dashed, line width=0.4pt] (4.277,-3.470) -- (4.277,0.200);
  \node[anchor=north, font=\sffamily\tiny, text=sciDark, inner sep=1pt] at (4.277,-3.510) {$-$5};
  \draw[sciGrid, dashed, line width=0.4pt] (4.686,-3.470) -- (4.686,0.200);
  \draw[sciGrid, dashed, line width=0.4pt] (5.094,-3.470) -- (5.094,0.200);
  \node[anchor=north, font=\sffamily\tiny, text=sciDark, inner sep=1pt] at (5.094,-3.510) {0};
  \draw[sciGrid, dashed, line width=0.4pt] (5.503,-3.470) -- (5.503,0.200);
  \draw[sciGrid, dashed, line width=0.4pt] (5.912,-3.470) -- (5.912,0.200);
  \node[anchor=north, font=\sffamily\tiny, text=sciDark, inner sep=1pt] at (5.912,-3.510) {+5};
  \draw[sciGrid, dashed, line width=0.4pt] (6.321,-3.470) -- (6.321,0.200);
  \draw[sciGrid, dashed, line width=0.4pt] (6.730,-3.470) -- (6.730,0.200);
  \node[anchor=north, font=\sffamily\tiny, text=sciDark, inner sep=1pt] at (6.730,-3.510) {+10};
  \draw[sciDark, line width=0.5pt] (5.094,-3.470) -- (5.094,0.200);
  \draw[sciDark, line width=0.5pt] (3.050,-3.470) -- (6.730,-3.470);
  \node[anchor=south, font=\sffamily\tiny, text=sciMuted, inner sep=1pt] at (5.094,0.200) {= GRPO};
  \node[anchor=north, font=\sffamily\scriptsize\bfseries, text=sciDark, inner sep=1pt] at (4.890,-3.820) {Ablation $-$ GRPO (pp)};
  \node[anchor=south west, font=\sffamily\tiny\bfseries, text=sciMuted, inner sep=1pt] at (0.000,0.160) {MoralChoice};
  \node[anchor=east, font=\sffamily\scriptsize, text=sciDark, inner sep=1pt] at (2.620,0.000) {Fragility, high amb.};
  \node[anchor=center, font=\sffamily\scriptsize, text=sciMuted, inner sep=1pt] at (2.740,0.000) {$\downarrow$};
  \node[anchor=east, font=\sffamily\scriptsize, text=sciDark, inner sep=1pt] at (2.620,-0.270) {Ambiguity recognized};
  \node[anchor=center, font=\sffamily\scriptsize, text=sciMuted, inner sep=1pt] at (2.740,-0.270) {$\uparrow$};
  \node[anchor=east, font=\sffamily\scriptsize, text=sciDark, inner sep=1pt] at (2.620,-0.540) {Fragility, low amb.};
  \node[anchor=center, font=\sffamily\scriptsize, text=sciMuted, inner sep=1pt] at (2.740,-0.540) {$\downarrow$};
  \node[anchor=south west, font=\sffamily\tiny\bfseries, text=sciMuted, inner sep=1pt] at (0.000,-0.950) {ELEPHANT};
  \node[anchor=east, font=\sffamily\scriptsize, text=sciDark, inner sep=1pt] at (2.620,-1.110) {Sycophancy};
  \node[anchor=center, font=\sffamily\scriptsize, text=sciMuted, inner sep=1pt] at (2.740,-1.110) {$\downarrow$};
  \node[anchor=east, font=\sffamily\scriptsize, text=sciDark, inner sep=1pt] at (2.620,-1.510) {Validation$^\ddagger$};
  \node[anchor=center, font=\sffamily\scriptsize, text=sciMuted, inner sep=1pt] at (2.740,-1.510) {$\downarrow$};
  \node[anchor=east, font=\sffamily\scriptsize, text=sciDark, inner sep=1pt] at (2.620,-2.040) {Indirectness$^\ddagger$};
  \node[anchor=center, font=\sffamily\scriptsize, text=sciMuted, inner sep=1pt] at (2.740,-2.040) {$\downarrow$};
  \node[anchor=east, font=\sffamily\scriptsize, text=sciDark, inner sep=1pt] at (2.620,-2.570) {Framing acceptance$^\ddagger$};
  \node[anchor=center, font=\sffamily\scriptsize, text=sciMuted, inner sep=1pt] at (2.740,-2.570) {$\downarrow$};
  \node[anchor=south west, font=\sffamily\tiny\bfseries, text=sciMuted, inner sep=1pt] at (0.000,-3.110) {MORABLES};
  \node[anchor=east, font=\sffamily\scriptsize, text=sciDark, inner sep=1pt] at (2.620,-3.270) {Accuracy};
  \node[anchor=center, font=\sffamily\scriptsize, text=sciMuted, inner sep=1pt] at (2.740,-3.270) {$\uparrow$};
  \draw[sciAmber, line width=0.7pt] (4.878,0.060) -- (5.551,0.060);
  \draw[sciAmber, line width=0.6pt] (4.878,0.025) -- (4.878,0.095);
  \draw[sciAmber, line width=0.6pt] (5.551,0.025) -- (5.551,0.095);
  \filldraw[fill=white, draw=sciAmber, line width=0.6pt] (5.138,0.060) -- (5.215,0.137) -- (5.292,0.060) -- (5.215,-0.017) -- cycle;
  \draw[sciTeal, line width=0.7pt] (4.950,-0.060) -- (5.600,-0.060);
  \draw[sciTeal, line width=0.6pt] (4.950,-0.095) -- (4.950,-0.025);
  \draw[sciTeal, line width=0.6pt] (5.600,-0.095) -- (5.600,-0.025);
  \filldraw[fill=white, draw=sciTeal, line width=0.6pt] (5.186,-0.122) -- (5.340,-0.122) -- (5.263,0.017) -- cycle;
  \draw[sciAmber, line width=0.7pt] (4.758,-0.210) -- (5.575,-0.210);
  \draw[sciAmber, line width=0.6pt] (4.758,-0.245) -- (4.758,-0.175);
  \draw[sciAmber, line width=0.6pt] (5.575,-0.245) -- (5.575,-0.175);
  \filldraw[fill=white, draw=sciAmber, line width=0.6pt] (5.090,-0.210) -- (5.167,-0.133) -- (5.243,-0.210) -- (5.167,-0.287) -- cycle;
  \draw[sciTeal, line width=0.7pt] (4.565,-0.330) -- (5.383,-0.330);
  \draw[sciTeal, line width=0.6pt] (4.565,-0.365) -- (4.565,-0.295);
  \draw[sciTeal, line width=0.6pt] (5.383,-0.365) -- (5.383,-0.295);
  \filldraw[fill=white, draw=sciTeal, line width=0.6pt] (4.897,-0.392) -- (5.051,-0.392) -- (4.974,-0.253) -- cycle;
  \draw[sciAmber, line width=0.7pt] (5.039,-0.480) -- (5.234,-0.480);
  \draw[sciAmber, line width=0.6pt] (5.039,-0.515) -- (5.039,-0.445);
  \draw[sciAmber, line width=0.6pt] (5.234,-0.515) -- (5.234,-0.445);
  \filldraw[fill=white, draw=sciAmber, line width=0.6pt] (5.045,-0.480) -- (5.122,-0.403) -- (5.199,-0.480) -- (5.122,-0.557) -- cycle;
  \draw[sciTeal, line width=0.7pt] (5.094,-0.600) -- (5.178,-0.600);
  \draw[sciTeal, line width=0.6pt] (5.094,-0.635) -- (5.094,-0.565);
  \draw[sciTeal, line width=0.6pt] (5.178,-0.635) -- (5.178,-0.565);
  \filldraw[fill=white, draw=sciTeal, line width=0.6pt] (5.045,-0.662) -- (5.199,-0.662) -- (5.122,-0.523) -- cycle;
  \draw[sciAmber, line width=0.7pt] (5.295,-1.050) -- (6.063,-1.050);
  \draw[sciAmber, line width=0.6pt] (5.295,-1.085) -- (5.295,-1.015);
  \draw[sciAmber, line width=0.6pt] (6.063,-1.085) -- (6.063,-1.015);
  \fill[sciAmber] (5.620,-1.050) -- (5.697,-0.973) -- (5.774,-1.050) -- (5.697,-1.127) -- cycle;
  \draw[sciTeal, line width=0.7pt] (5.519,-1.170) -- (6.215,-1.170);
  \draw[sciTeal, line width=0.6pt] (5.519,-1.205) -- (5.519,-1.135);
  \draw[sciTeal, line width=0.6pt] (6.215,-1.205) -- (6.215,-1.135);
  \fill[sciTeal] (5.773,-1.232) -- (5.927,-1.232) -- (5.850,-1.093) -- cycle;
  \draw[sciAmber, line width=0.7pt] (4.601,-1.510) -- (4.953,-1.510);
  \draw[sciAmber, line width=0.6pt] (4.601,-1.545) -- (4.601,-1.475);
  \draw[sciAmber, line width=0.6pt] (4.953,-1.545) -- (4.953,-1.475);
  \fill[sciAmber] (4.700,-1.510) -- (4.777,-1.433) -- (4.854,-1.510) -- (4.777,-1.587) -- cycle;
  \draw[sciTeal, line width=0.7pt] (4.989,-1.670) -- (5.236,-1.670);
  \draw[sciTeal, line width=0.6pt] (4.989,-1.705) -- (4.989,-1.635);
  \draw[sciTeal, line width=0.6pt] (5.236,-1.705) -- (5.236,-1.635);
  \filldraw[fill=white, draw=sciTeal, line width=0.6pt] (5.029,-1.732) -- (5.183,-1.732) -- (5.106,-1.593) -- cycle;
  \draw[sciNavy, line width=0.7pt] (3.378,-1.360) -- (4.119,-1.360);
  \draw[sciNavy, line width=0.6pt] (3.378,-1.395) -- (3.378,-1.325);
  \draw[sciNavy, line width=0.6pt] (4.119,-1.395) -- (4.119,-1.325);
  \filldraw[fill=sciNavy, draw=sciNavy, line width=0.6pt] (3.689,-1.425) rectangle (3.819,-1.295);
  \draw[sciAmber, line width=0.7pt] (4.342,-2.040) -- (4.812,-2.040);
  \draw[sciAmber, line width=0.6pt] (4.342,-2.075) -- (4.342,-2.005);
  \draw[sciAmber, line width=0.6pt] (4.812,-2.075) -- (4.812,-2.005);
  \fill[sciAmber] (4.500,-2.040) -- (4.577,-1.963) -- (4.654,-2.040) -- (4.577,-2.117) -- cycle;
  \draw[sciTeal, line width=0.7pt] (5.694,-2.200) -- (6.341,-2.200);
  \draw[sciTeal, line width=0.6pt] (5.694,-2.235) -- (5.694,-2.165);
  \draw[sciTeal, line width=0.6pt] (6.341,-2.235) -- (6.341,-2.165);
  \fill[sciTeal] (5.935,-2.262) -- (6.088,-2.262) -- (6.012,-2.123) -- cycle;
  \draw[sciNavy, line width=0.7pt] (4.965,-1.890) -- (5.130,-1.890);
  \draw[sciNavy, line width=0.6pt] (4.965,-1.925) -- (4.965,-1.855);
  \draw[sciNavy, line width=0.6pt] (5.130,-1.925) -- (5.130,-1.855);
  \filldraw[fill=white, draw=sciNavy, line width=0.6pt] (4.982,-1.955) rectangle (5.112,-1.825);
  \draw[sciAmber, line width=0.7pt] (5.706,-2.570) -- (6.376,-2.570);
  \draw[sciAmber, line width=0.6pt] (5.706,-2.605) -- (5.706,-2.535);
  \draw[sciAmber, line width=0.6pt] (6.376,-2.605) -- (6.376,-2.535);
  \fill[sciAmber] (5.970,-2.570) -- (6.047,-2.493) -- (6.124,-2.570) -- (6.047,-2.647) -- cycle;
  \draw[sciTeal, line width=0.7pt] (5.071,-2.730) -- (5.635,-2.730);
  \draw[sciTeal, line width=0.6pt] (5.071,-2.765) -- (5.071,-2.695);
  \draw[sciTeal, line width=0.6pt] (5.635,-2.765) -- (5.635,-2.695);
  \filldraw[fill=white, draw=sciTeal, line width=0.6pt] (5.264,-2.792) -- (5.418,-2.792) -- (5.341,-2.653) -- cycle;
  \draw[sciNavy, line width=0.7pt] (5.716,-2.420) -- (6.624,-2.420);
  \draw[sciNavy, line width=0.6pt] (5.716,-2.455) -- (5.716,-2.385);
  \draw[sciNavy, line width=0.6pt] (6.624,-2.455) -- (6.624,-2.385);
  \filldraw[fill=sciNavy, draw=sciNavy, line width=0.6pt] (6.105,-2.485) rectangle (6.235,-2.355);
  \draw[sciAmber, line width=0.7pt] (5.202,-3.210) -- (5.687,-3.210);
  \draw[sciAmber, line width=0.6pt] (5.202,-3.245) -- (5.202,-3.175);
  \draw[sciAmber, line width=0.6pt] (5.687,-3.245) -- (5.687,-3.175);
  \fill[sciAmber] (5.364,-3.210) -- (5.440,-3.133) -- (5.517,-3.210) -- (5.440,-3.287) -- cycle;
  \draw[sciTeal, line width=0.7pt] (5.071,-3.330) -- (5.610,-3.330);
  \draw[sciTeal, line width=0.6pt] (5.071,-3.365) -- (5.071,-3.295);
  \draw[sciTeal, line width=0.6pt] (5.610,-3.365) -- (5.610,-3.295);
  \filldraw[fill=white, draw=sciTeal, line width=0.6pt] (5.264,-3.392) -- (5.417,-3.392) -- (5.341,-3.253) -- cycle;
  \draw[sciAmber, line width=0.7pt] (-0.270,-4.330) -- (0.030,-4.330);
  \draw[sciAmber, line width=0.6pt] (-0.270,-4.365) -- (-0.270,-4.295);
  \draw[sciAmber, line width=0.6pt] (0.030,-4.365) -- (0.030,-4.295);
  \fill[sciAmber] (-0.197,-4.330) -- (-0.120,-4.253) -- (-0.043,-4.330) -- (-0.120,-4.407) -- cycle;
  \node[anchor=west, font=\sffamily\tiny, text=sciDark, inner sep=1pt, text height=1.5ex, text depth=0.25ex] at (0.080,-4.330) {Unmasked (masking)};
  \draw[sciTeal, line width=0.7pt] (2.220,-4.330) -- (2.520,-4.330);
  \draw[sciTeal, line width=0.6pt] (2.220,-4.365) -- (2.220,-4.295);
  \draw[sciTeal, line width=0.6pt] (2.520,-4.365) -- (2.520,-4.295);
  \fill[sciTeal] (2.293,-4.392) -- (2.447,-4.392) -- (2.370,-4.253) -- cycle;
  \node[anchor=west, font=\sffamily\tiny, text=sciDark, inner sep=1pt, text height=1.5ex, text depth=0.25ex] at (2.570,-4.330) {Synthetic (curriculum)};
  \draw[sciNavy, line width=0.7pt] (4.810,-4.330) -- (5.110,-4.330);
  \draw[sciNavy, line width=0.6pt] (4.810,-4.365) -- (4.810,-4.295);
  \draw[sciNavy, line width=0.6pt] (5.110,-4.365) -- (5.110,-4.295);
  \filldraw[fill=sciNavy, draw=sciNavy, line width=0.6pt] (4.895,-4.395) rectangle (5.025,-4.265);
  \node[anchor=west, font=\sffamily\tiny, text=sciDark, inner sep=1pt, text height=1.5ex, text depth=0.25ex] at (5.160,-4.330) {31B judge (Unmasked)};
\end{tikzpicture}%
  \caption{Held-out benchmark change, Unmasked and Synthetic vs.\ GRPO (percentage points; item-paired 95\% bootstrap CIs; rows as in Figure~\ref{fig:benchmark_change}). Squares: the same Unmasked$-$GRPO difference re-scored by a frozen 31B judge (Synthetic not re-scored). Hollow: CI includes zero.}
  \label{fig:ablation_effects}
\end{figure}
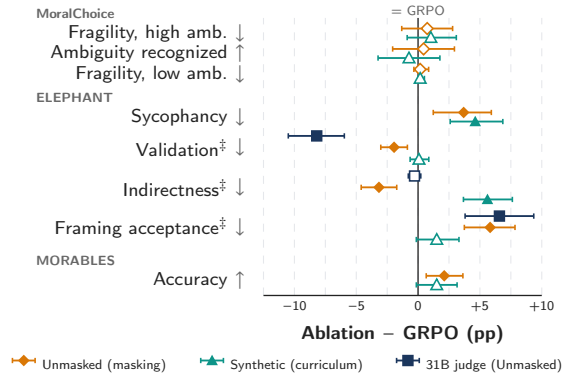

RQ3 asks whether this divergence depends on loss masking or narrative grounding (Figure~\ref{fig:ablation_effects}). With one seed per run, we attribute an effect to a factor only if one ablation diverges from GRPO and the other does not. The divergence itself recurs in all three runs. Both ablations keep GRPO's consistency under reordering, changing the presentation-flip rate by only \benchMcHighFragileUMinusG{} points in Unmasked (95\% CI \benchMcHighFragileUMinusGCI{}) and \benchMcHighFragileSMinusG{} in Synthetic (\benchMcHighFragileSMinusGCI{}), yet both are more sycophantic than GRPO (\benchElSycophancyUMinusG{} points, 95\% CI \benchElSycophancyUMinusGCI{}; \benchElSycophancySMinusG{}, \benchElSycophancySMinusGCI{}).

Masking decides where the policy spends its words. The training reward favors longer thinking in all runs (\selThinkG{} per SD in GRPO, \selThinkU{} in Unmasked; Figure~\ref{fig:reward_selection}), but masking hides thinking tokens from the policy gradient. GRPO's thinking shrinks from \allocThinkStart{} to \allocThinkEnd{} words while its answers grow from \allocRespStart{} to \allocRespEnd{} (Figure~\ref{fig:drift_dynamics}). Unmasked follows the reward instead: its thinking grows (\allocUThinkPct{}) more than its answers (\allocURespPct{}), whose growth stalls in epoch~2 (\allocUEpTwoShare{} of the total, vs.\ \allocEpTwoShare{} in GRPO). Masking also reduces the lexical overlap between thinking and answers (\overlapG{}; Unmasked: \overlapU{}).

This reallocation trades one benchmark against another. On MORABLES fables, Unmasked recovers part of GRPO's accuracy loss (\benchMorMcqaUMinusG{} points over GRPO, 95\% CI \benchMorMcqaUMinusGCI{}) without thinking longer (\benchCotMorablesUnmasked{} vs.\ \benchCotMorablesGRPO{} words; Figure~\ref{fig:benchmark_cot_tokens} in Appendix~\ref{sec:app-cot-benchmark-length}), but stays below Base. On ELEPHANT, it thinks longer than GRPO (\benchCotElephantUnmasked{} vs.\ \benchCotElephantGRPO{} words) and accepts the user's framing more often (\benchElFramingUMinusG{} points, 95\% CI \benchElFramingUMinusGCI{}; 31B judge: \benchElFramingExtUMinusG{}, \benchElFramingExtUMinusGCI{}). Its self-scored gain in directness (\benchElIndirectnessUMinusG{} points, 95\% CI \benchElIndirectnessUMinusGCI{}) does not survive the 31B judge, under which both runs are already near zero.

The ablations differ in how much register they carry. Of the features GRPO shifts significantly ($q < .05$), Synthetic replicates \replShareS{} at nearly full strength (\replRatioS{} median magnitude). Unmasked replicates a similar share (\replShareU{}) but at only \replRatioU{} of GRPO's magnitude. Literature thus barely shapes the register, whereas unmasking dilutes it.

\section{Discussion}
\label{sec:discussion}

The goal of self-improvement for machine ethics is normative competence, not persuasiveness~\citep{talat2022machine, hancox-li2024ethics, sachdeva2025normative}. Our 4B policy learns a prescriptive register that transfers as decisiveness rather than as better judgment.

\subsection{The Judge Selects the Register}
\label{sec:discussion-eloquence}

Optimizing against an automated judge invites Goodhart's law: a policy can satisfy the judge through whatever correlates with its preference~\citep{gao2023scaling, zheng2023judging}. The change we observe is not a set of disparate surface tricks but one register that recurs under every assigned framework (\S\ref{sec:results-rq1}).

Judge validation raised a specific risk: on human-written arguments, the 4B judge's overall preference agreed most often with its rhetoric preference, its weakest predictor of human preference (\S\ref{sec:judgeval}). On the policy's own completions, however, diagnostic logging across \judgeGroups{} prompt groups shows that the overall reward follows dialectic (mean Spearman $\rho = \judgeAgreeDialectic{}$), then logic (\judgeAgreeLogic{}), and rhetoric least (\judgeAgreeRhetoric{}). Rhetoric still leaves a trace: it specifically favors shallower dependency trees (\judgeRhetDepth{}~SD per SD of maximum depth, 95\% CI \judgeRhetDepthCI{}). Outcome accuracy on human arguments therefore does not reveal which aspect drives the judge on the policy's distribution \citep{wang2026outcome}; only the policy's own rollouts do.

The answers grow although the training reward does not favor length. The raw judge favors longer answers within each group (\judgeResp{}~SD of advantage per SD of answer length, 95\% CI \judgeRespCI{}), but the leave-one-out length penalty flattens this preference (\selRespG{}~SD). Answers lengthen nonetheless, because within-group normalization cannot penalize drift that all completions in a group share. The register itself is one source of such drift: in the policy's output, the rewarded stance markers co-occur with longer answers, and masking leaves the visible answer as the only channel for optimization pressure. GRPO's response-level loss normalization may also favor longer completions~\citep{liu2025understanding}. Either way, the growth is a byproduct of learning the register under masking, not length gaming against a slipping penalty.
\subsection{Decisive but Pliable}
\label{sec:discussion-seduction}

The policy commits more often and more stably, but it commits within the frame the user supplies. Consistency under reordering and framing acceptance rise in GRPO, neither ablation reverses them, and sycophancy falls in no run. This consistency is specific to the rewarded role: the same weights that halve presentation flips on MoralChoice flip more often as a pairwise judge on IBM-ArgQ (\judgeIbmFourBFlipRate{} before training, \judgeIbmGrpoFlipRate{} after; \S\ref{sec:judgeval}). One account links decisiveness and pliability: a policy rewarded for verdicts learns to reach one, and the cheapest premise for a verdict is the one in the prompt. Normative ethics offers no rule-based verifier that would penalize a false premise~\citep{creswell2022faithful, jacovi2024chain-of-thought}. In-context demonstrations fit the same account: they make Base the most direct and most consistent model under reordering we test, and they raise framing acceptance more than any training run does. The rise in framing acceptance is not an artifact of self-scoring: a frozen 31B judge registers it in GRPO and, more strongly, in Unmasked. Persistent sycophancy is not harmless: even one sycophantic exchange makes users less willing to repair conflicts \citep{cheng2026sycophantic}.

\subsection{Masking Is a Lever with a Price}
\label{sec:discussion-masking}

Masking determines which channel absorbs policy optimization (\S\ref{sec:results-rq3}). Without masking, the policy gradient spreads across the full sequence instead of concentrating on the answer; under per-token loss normalization, this dilution may explain why Unmasked acquires the register more weakly and degrades less on MORABLES fables.

This choice creates a dilemma for chain-of-thought monitorability~\citep{korbak2025chain}. No policy gradient reaches thinking tokens, yet the answer repeats less of the thinking (\S\ref{sec:results-rq3}), consistent with feedback spillover through shared weights \citep{drori2026output}. Lexical overlap is not faithfulness~\citep{jacovi2024chain-of-thought}, but masking risks divorcing the stated justification from the verdict.

Exposing thinking to optimization does not resolve the policy's pliability either. Unmasked thinks longer than GRPO on ELEPHANT and accepts the user's framing more often, yet longer traces go with framing acceptance only within GRPO (Table~\ref{tab:elephant_framing_length} in Appendix~\ref{sec:app-cot-benchmark-length}). Without an external verifier, longer deliberation does not make the policy judge the user's premises independently.

\subsection{What Narrative Grounding Changes}
\label{sec:discussion-curriculum}

Grounding training dilemmas in classical literature does not induce the prescriptive register: Synthetic replicates most of it at nearly full strength (\S\ref{sec:results-rq3}), so the register reflects the automated judge's preferences rather than the stylistic richness of nineteenth-century texts. Narrative grounding reliably influences only directness: synthetic training makes answers \benchElIndirectnessSMinusG{} points more indirect (95\% CI \benchElIndirectnessSMinusGCI{}). Verified literary conflicts ensure genuine moral tension during training, but pairwise reinforcement learning distills a way of arguing rather than sound moral deliberation.

\section{Conclusion}
\label{sec:conclusion}

We investigated whether rewarding argument quality in a self-judged tournament improves moral judgment. We trained a Gemma-4 4B policy with masked thinking on literature dilemmas verified with an SMT solver. Our judge rates argument quality like human annotators, and reinforcement learning instills a prescriptive register: answers trade caveats and backward justifications for forward obligations across every assigned ethical framework. On held-out benchmarks, this register makes the policy decisive, halving presentation flips on ambiguous MoralChoice scenarios and making its ELEPHANT replies more direct.

Decisive verdicts, however, do not yield more valid moral judgments: on ELEPHANT posts, moral sycophancy remains flat while framing acceptance rises; on MORABLES fables, allegorical accuracy falls. Both ablations match the reordering gains but show a weaker register shift and higher moral sycophancy. Training on synthetic dilemmas preserves most of the register, so it comes from the reward rather than the literary source. Unmasking the thinking weakens the register most and reduces the MORABLES drop, but increases framing acceptance.
Consistency under reordering, a common proxy for moral validity, can thus be optimized without improving the moral judgment it is meant to signal. Machine ethics research should therefore evaluate consistency and validity separately.

\section*{Limitations}
\label{sec:limitations}

Our reward comes from a single design: the policy judges its own completions with shared weights. Because no run trains against a fixed judge such as the 31B model, we cannot tell whether the register arises from self-judgment or from pairwise argument-quality rewards in general. Shared weights may also lead the judge to favor familiar completions, a form of self-preference bias~\citep{panickssery2024llm}. We have also not tested the prediction of \S\ref{sec:discussion-seduction} that framing acceptance rises with decisiveness during training.

Our judges are validated only on human-written arguments (\S\ref{sec:judgeval}), not on policy completions, and no human evaluates the policy's arguments. The central divergence therefore rests on three metrics that need no judge: presentation flips, moral sycophancy, and MORABLES accuracy. The evaluated model and a frozen Gemma-4 31B judge from the same model family score the remaining ELEPHANT metrics, and neither is validated against human annotations. The two judges agree on the direction of change but only weakly on individual items (Appendix~\ref{sec:app-icl-campaign}), so we interpret direction rather than level.

Our findings come from a narrow experimental scope. We train one policy scale, Gemma-4 4B, always with thinking enabled, and each configuration only once. The divergence recurs in all three runs (\S\ref{sec:results-rq3}), but our confidence intervals reflect variation across dilemmas and benchmark items, not across training runs. All dilemmas are in English, and the literature dilemmas come from four nineteenth-century Western novels.

Formal verification covers the conflict, not its resolution. Z3 checks the extractor's Boolean encoding of each dilemma, not the narrative itself, and the encoding states every duty as a hard constraint (\S\ref{sec:extraction}). The check confirms that no action fulfills both duties, but the encoding cannot express that one duty outweighs the other. Many dilemmas turn on such an ordering~\citep{bench-capon2003persuasion}, for example whether preventing serious harm outweighs keeping a promise. Resolving a dilemma is thus left to the judge, which no formal check constrains. Weighted encodings, such as soft constraints in MaxSMT, could make these orderings explicit.

\section*{Ethical Considerations}
\label{sec:ethics}

The main risk of this work follows from its main finding. Rewarding argument quality produced a model that commits to verdicts more readily and more stably. The same model accepts the user's framing more often, and its moral judgment does not improve: MORABLES accuracy falls. Deployed as a moral advisor, such a model would lend confident language to whatever premises a user brings, and sycophantic advice already lowers users' willingness to repair conflicts \citep{cheng2026sycophantic}. Because the policy becomes more consistent under reordering, consistency checks would hide this risk. Decisiveness and consistency are therefore insufficient evidence of valid moral judgment. The same reward also teaches the policy to argue any assigned framework in one confident register (\S\ref{sec:results-rq1}), so the method could produce persuasive rationalizations for a predetermined verdict. We do not release the trained models.

We take no position on which ethical framework is correct. The five frameworks serve as assigned lenses, and the policy's verdicts are not moral ground truth. The benchmarks carry their own normative assumptions: ELEPHANT counts accepting the user's framing as a failure, and MORABLES scores each fable against a single reference moral. All five frameworks come from the Western philosophical tradition, which leaves non-Western and communal ethics unrepresented~\citep{romberg2025towards}. The source novels also reflect the views of their period on class, gender, and religion, and the extracted dilemmas may carry these views.

We collected no new data from human subjects. The training texts are public-domain books from PG19, and all models, benchmarks, and datasets are used in accordance with their licenses and terms of use. ELEPHANT contains posts written by online users; we use the released data as provided and make no attempt to identify their authors. We used AI assistants for writing and coding; the authors reviewed all output and take full responsibility for the content.

\bibliography{paper}

\appendix
\setcounter{topnumber}{3}%
\setcounter{totalnumber}{4}%
\renewcommand{\topfraction}{0.9}%
\renewcommand{\textfraction}{0.1}%
\renewcommand{\floatpagefraction}{0.8}%

\section{Dilemma Data}
\label{sec:app-grounding}

\subsection{Literature Pool}
\label{sec:app-datasheet}

The literature pool (\textsc{LitPool}) is drawn from four PG19 novels by Hugo, Austen, Melville, and Dostoevsky. \emph{Les Mis\'{e}rables} supplies the largest share of the extracted dilemmas, and compassion is the most frequent competing value (Table~\ref{tab:datasheet}). The extraction prompt is shown in Appendix~\ref{sec:app-extraction-prompt}.

\begin{table}[t]
  \centering
  \small
  \setlength{\tabcolsep}{4pt}
  \begin{tabular}{llr}
    \toprule
    Author & Source Work & Extracted \\
    \midrule
    Hugo & \emph{Les Mis\'{e}rables} (\#135) & 88 \\
    Austen & \emph{Pride and Prejudice} (\#1342) & 39 \\
    Melville & \emph{Moby-Dick} (\#2701) & 33 \\
    Dostoevsky & \emph{Crime and Punishment} (\#2554) & 25 \\
    \midrule
    \textbf{Total} & & \textbf{185} \\
    \bottomrule
  \end{tabular}

  \vspace{1ex}

  \begin{tabular}{lrr}
    \toprule
    Value-concept & Count & \% of pool \\
    \midrule
    compassion & 41 & 22.2\% \\
    honesty & 19 & 10.3\% \\
    integrity & 18 & 9.7\% \\
    self preservation & 18 & 9.7\% \\
    professional duty & 15 & 8.1\% \\
    moral discipline & 14 & 7.6\% \\
    autonomy & 14 & 7.6\% \\
    familial duty & 11 & 5.9\% \\
    legal obedience & 9 & 4.9\% \\
    pragmatism & 7 & 3.8\% \\
    \bottomrule
  \end{tabular}
  \caption{Curriculum dilemma pool (\textsc{LitPool}). Top: dilemmas extracted per PG19 source work (Gutenberg ID). Bottom: most frequent competing value-concepts among the \poolNumExtracted{} extracted dilemmas, of which \poolNumTrained{} pass the Z3 gate and are trained on.}
  \label{tab:datasheet}
\end{table}

\subsection{Z3 Gate Failures}
\label{sec:app-zfail}

Most generated synthetic dilemmas pass the Z3 gate (\zfailNaiveZtwoPct{}; Table~\ref{tab:zfail}). Most of the rest are not genuine dilemmas: their duties can be satisfied together ($Z{=}1$, \zfailNaiveZonePct{}). The few degenerate encodings ($Z{=}0$) are mostly over-constrained: the premise fixes a fact that one duty then negates, so that duty alone is unsatisfiable. The Synthetic run trains on the gate passes of this batch and of a second batch, with exact duplicates removed.

\begin{table}[t]
  \centering
  \small
  \begin{tabular}{@{}lr@{}}
    \toprule
    Z3 gate outcome & Dilemmas \\
    \midrule
    Generated ($N$) & 130 \\
    \quad $Z{=}2$ (exclusive) & 101 (77.7\%) \\
    \quad $Z{=}1$ (co-satisfiable) & 20 (15.4\%) \\
    \quad $Z{=}0$ (degenerate) & 9 (6.9\%) \\
    \midrule
    \multicolumn{2}{@{}l}{\textit{$Z{=}0$ failure modes}} \\
    \quad One duty UNSAT & 8 \\
    \quad Premise UNSAT & 0 \\
    \quad SMT parse error & 1 \\
    \midrule
    Empty premise (within $Z{=}1$) & 7 \\
    \bottomrule
  \end{tabular}
  \caption{Outcomes of the Z3 gate on the first generation batch of the synthetic dilemmas the Synthetic run trains on. $Z{=}2$: the duties are mutually exclusive (a genuine dilemma). $Z{=}1$: the duties are co-satisfiable. $Z{=}0$: the encoding is degenerate.}
  \label{tab:zfail}
\end{table}

\subsection{Literature vs.\ Synthetic Dilemmas}
\label{sec:app-generality}

The literature and synthetic dilemmas span a similar number of semantic axes (Table~\ref{tab:generality}). We measure semantic diversity with the Vendi score \citep{friedman2023vendi} and the effective rank \citep{roy2007effective} on three embedders: bge \citep{xiao2024c-pack}, unixcoder \citep{guo2022unixcoder}, and gemini-embedding \citep{lee2025gemini}. The literature pool has the higher effective rank on all three embedders, but the jackknife CIs of the two pools overlap on every embedder (gemini-embedding: [\rqgErankGemLCIlo{}, \rqgErankGemLCIhi{}] vs.\ [\rqgErankGemSnaiveCIlo{}, \rqgErankGemSnaiveCIhi{}]).

\begin{table*}[t]
  \centering
  \small
  \begin{tabular}{lrrrrrrrrr}
    \toprule
    Source & $n$ & \multicolumn{3}{c}{Vendi $\uparrow$} & \multicolumn{3}{c}{Eff.\ rank $\uparrow$} & NN $\downarrow$ & Conc. \\
    \cmidrule(lr){3-5}\cmidrule(lr){6-8}
    & & bge & uc & gem & bge & uc & gem & bge & \\
    \midrule
    Hugo & 88 & 3.79 & 7.02 & 2.75 & 22.40 & 23.12 & 21.97 & 0.833 & 16 \\
    Dostoevsky & 25 & 3.89 & 6.67 & 2.71 & 22.02 & 22.96 & 21.22 & 0.839 & 20 \\
    Austen & 39 & 3.51 & 6.29 & 2.39 & 22.53 & 23.02 & 21.95 & 0.839 & 24 \\
    Melville & 33 & 4.41 & 6.62 & 3.10 & 22.71 & 23.10 & 22.31 & 0.795 & 25 \\
    \midrule
    \textsc{LitPool} & 185 & 4.39 & 6.97 & 3.27 & 22.63 & 23.19 & 22.20 & 0.806 & 25 \\
    Synthetic & 72 & 6.48 & 7.96 & 3.78 & 22.19 & 22.85 & 22.07 & 0.791 & 31 \\
    \bottomrule
  \end{tabular}
  \caption{Intrinsic diversity of the four-author literature pool (\textsc{LitPool}), its source works, and synthetic dilemmas from the generator of the Synthetic run, on three embedders (bge, unixcoder, gemini-embedding) at $n{=}25$ per source (seed 42). $n$: dilemmas available to this analysis. NN: mean nearest-neighbor cosine. Conc.: distinct value-concepts. Vendi is embedder-scaled: compare within a column.}
  \label{tab:generality}
\end{table*}

\section{Training Configuration}
\label{sec:appendix_training}

Table~\ref{tab:hyperparameters} lists the configuration of the GRPO run. We train all three runs with Tunix \citep{bao2025tunix}. Each training run (GRPO, Unmasked, Synthetic) took about 24 hours.

\begin{table}[t]
\centering
\small
\setlength{\tabcolsep}{4pt}
\begin{tabular}{@{}lp{4.3cm}@{}}
\toprule
Component & Setting \\
\midrule
\multicolumn{2}{@{}l}{\textit{Model and hardware}} \\
Policy & \raggedright Gemma-4 4B, instruction-tuned \tabularnewline
Hardware & \raggedright 32 accelerator chips, FSDP $16 \times$ TP $2$ \tabularnewline
Sequence length & \raggedright 1,024 prompt + 3,072 generation tokens \tabularnewline
\midrule
\multicolumn{2}{@{}l}{\textit{Data and schedule}} \\
Training data & \raggedright \poolNumTrained{} Z3-verified \textsc{LitPool} dilemmas, 2 epochs \tabularnewline
Frameworks & \raggedright 5 per dilemma (\S\ref{sec:frameworks}) \tabularnewline
Batch size & \raggedright 32 groups per step, $G{=}8$ completions each \tabularnewline
Mini-/micro-batch & \raggedright 32 / 2 \tabularnewline
Rollout sampling & \raggedright $T{=}1.0$, top-$p{=}1.0$, top-$k{=}64$ \tabularnewline
\midrule
\multicolumn{2}{@{}l}{\textit{Optimization}} \\
Optimizer & \raggedright AdamW ($\beta_1{=}0.9$, $\beta_2{=}0.99$, weight decay 0.01) \tabularnewline
Learning rate & \raggedright $10^{-5}$ peak, 19\% warmup, cosine decay to $10^{-6}$ \tabularnewline
Gradient clipping & \raggedright 1.0 (max.\ norm) \tabularnewline
PPO clip ratio $\epsilon$ & \raggedright 0.2 \tabularnewline
KL penalty $\beta$ & \raggedright 0.02 (to the reference model) \tabularnewline
Advantage & \raggedright RLOO, unnormalized (Eq.~\ref{eq:rloo}) \tabularnewline
Loss masking & \raggedright No gradient on thinking tokens (\S\ref{sec:reward}) \tabularnewline
\midrule
\multicolumn{2}{@{}l}{\textit{Self-judge and reward}} \\
Judge & \raggedright The policy itself (shared weights) \tabularnewline
Judge decoding & \raggedright Thinking on, at most 3,072 tokens; $T{=}1.0$, top-$p{=}0.95$, top-$k{=}64$ \tabularnewline
Tournament & \raggedright Round robin, $\binom{8}{2}{=}28$ pairs in both orders (56 judgments) \tabularnewline
Presentation flip & \raggedright Scored as a tie \tabularnewline
Preference model & \raggedright Rao--Kupper with ties ($\nu > 1$) \tabularnewline
MAP prior & \raggedright Symmetric neutral baseline ($\alpha{=}1.0$) \tabularnewline
Length penalty & \raggedright Within-group $z$-score, $\lambda_{\mathrm{len}}{=}\ablLenPenaltyCoeff{}$ \tabularnewline
\bottomrule
\end{tabular}
\caption{Training configuration of the GRPO run.}
\label{tab:hyperparameters}
\end{table}

\section{Judge Rubric and Validation}
\label{sec:appendix_rubric}

The judge prompt follows the taxonomy of \citet{wachsmuth2017computational}: three aspects, each with an aspect-level dimension and its sub-dimensions, plus overall quality (Table~\ref{tab:wachsmuth_rubric}). The judge reasons through all 15 dimensions in its thinking and then emits four verdicts, one per aspect and one for overall quality. Only the overall verdict enters the reward.

\begin{table}[t]
  \centering
  \small
  \setlength{\tabcolsep}{4pt}
  \begin{tabular}{@{}lp{4.4cm}@{}}
    \toprule
    Dimension & Question in the judge prompt \\
    \midrule
    \multicolumn{2}{@{}l}{\textit{Logic}} \\
    Cogency & \raggedright Are the premises acceptable, relevant to the conclusion, and sufficient to draw it? \tabularnewline
    \quad Local acceptability & \raggedright Are the premises rationally worthy of belief? \tabularnewline
    \quad Local relevance & \raggedright Do the premises actually bear on the conclusion? \tabularnewline
    \quad Local sufficiency & \raggedright Do the premises together give enough support? \tabularnewline
    \midrule
    \multicolumn{2}{@{}l}{\textit{Rhetoric}} \\
    Effectiveness & \raggedright Does the argument persuade the target audience of the stance? \tabularnewline
    \quad Credibility & \raggedright Does it make the author worthy of credence (ethos)? \tabularnewline
    \quad Emotional appeal & \raggedright Does it make a successful emotional appeal (pathos)? \tabularnewline
    \quad Clarity & \raggedright Does it use correct, widely unambiguous, on-topic language? \tabularnewline
    \quad Appropriateness & \raggedright Are the language and tone proportional to the issue? \tabularnewline
    \quad Arrangement & \raggedright Are the issue, arguments, and conclusion in a proper order? \tabularnewline
    \midrule
    \multicolumn{2}{@{}l}{\textit{Dialectic}} \\
    Reasonableness & \raggedright Does it help resolve the issue acceptably and sufficiently? \tabularnewline
    \quad Global acceptability & \raggedright Would the audience accept the arguments and framing? \tabularnewline
    \quad Global relevance & \raggedright Does it contribute to resolving the issue at hand? \tabularnewline
    \quad Global sufficiency & \raggedright Does it rebut anticipatable counter-arguments? \tabularnewline
    \midrule
    Overall quality & \raggedright A holistic judgment of overall argument quality, considering all dimensions above. \tabularnewline
    \bottomrule
  \end{tabular}
  \caption{The 15 argument-quality dimensions of \citet{wachsmuth2017computational} as phrased in the judge prompt (Appendix~\ref{sec:appendix_prompt}), grouped by aspect; sub-dimensions are indented under their aspect. The judge reasons over all 15 but emits only four verdicts: logic, rhetoric, dialectic, and overall quality.}
  \label{tab:wachsmuth_rubric}
\end{table}

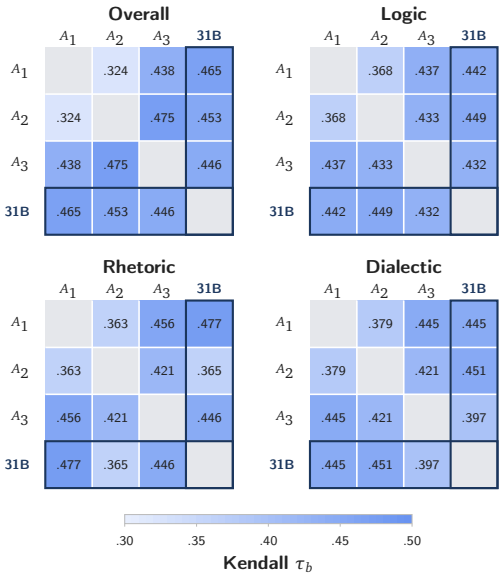
\begin{figure}[t]
  \centering
\providecolor{sciNavy}{HTML}{1B365D}%
\providecolor{sciBlue}{HTML}{2563EB}%
\providecolor{sciDark}{HTML}{222222}%
\providecolor{sciMuted}{HTML}{666666}%
\providecolor{sciGrid}{HTML}{E5E7EB}%
\begin{tikzpicture}[x=1cm, y=1cm]%
  \node[anchor=base, font=\scriptsize\sffamily\bfseries, text=sciDark] at (1.790, 7.110) {Overall};
  \node[anchor=base, font=\tiny\sffamily, text=sciDark] at (0.860, 6.850) {$A_1$};
  \node[anchor=east, font=\tiny\sffamily, text=sciDark] at (0.470, 6.440) {$A_1$};
  \node[anchor=base, font=\tiny\sffamily, text=sciDark] at (1.480, 6.850) {$A_2$};
  \node[anchor=east, font=\tiny\sffamily, text=sciDark] at (0.470, 5.820) {$A_2$};
  \node[anchor=base, font=\tiny\sffamily, text=sciDark] at (2.100, 6.850) {$A_3$};
  \node[anchor=east, font=\tiny\sffamily, text=sciDark] at (0.470, 5.200) {$A_3$};
  \node[anchor=base, font=\tiny\sffamily, text=sciNavy] at (2.720, 6.850) {\textbf{31B}};
  \node[anchor=east, font=\tiny\sffamily, text=sciNavy] at (0.470, 4.580) {\textbf{31B}};
  \fill[sciGrid] (0.550, 6.130) rectangle (1.170, 6.750);
  \fill[sciBlue!17] (1.170, 6.130) rectangle (1.790, 6.750);
  \node[font=\tiny\sffamily, text=sciDark] at (1.480, 6.440) {.324};
  \fill[sciBlue!51] (1.790, 6.130) rectangle (2.410, 6.750);
  \node[font=\tiny\sffamily, text=sciDark] at (2.100, 6.440) {.438};
  \fill[sciBlue!60] (2.410, 6.130) rectangle (3.030, 6.750);
  \node[font=\tiny\sffamily, text=sciDark] at (2.720, 6.440) {.465};
  \fill[sciBlue!17] (0.550, 5.510) rectangle (1.170, 6.130);
  \node[font=\tiny\sffamily, text=sciDark] at (0.860, 5.820) {.324};
  \fill[sciGrid] (1.170, 5.510) rectangle (1.790, 6.130);
  \fill[sciBlue!62] (1.790, 5.510) rectangle (2.410, 6.130);
  \node[font=\tiny\sffamily, text=sciDark] at (2.100, 5.820) {.475};
  \fill[sciBlue!56] (2.410, 5.510) rectangle (3.030, 6.130);
  \node[font=\tiny\sffamily, text=sciDark] at (2.720, 5.820) {.453};
  \fill[sciBlue!51] (0.550, 4.890) rectangle (1.170, 5.510);
  \node[font=\tiny\sffamily, text=sciDark] at (0.860, 5.200) {.438};
  \fill[sciBlue!62] (1.170, 4.890) rectangle (1.790, 5.510);
  \node[font=\tiny\sffamily, text=sciDark] at (1.480, 5.200) {.475};
  \fill[sciGrid] (1.790, 4.890) rectangle (2.410, 5.510);
  \fill[sciBlue!54] (2.410, 4.890) rectangle (3.030, 5.510);
  \node[font=\tiny\sffamily, text=sciDark] at (2.720, 5.200) {.446};
  \fill[sciBlue!60] (0.550, 4.270) rectangle (1.170, 4.890);
  \node[font=\tiny\sffamily, text=sciDark] at (0.860, 4.580) {.465};
  \fill[sciBlue!56] (1.170, 4.270) rectangle (1.790, 4.890);
  \node[font=\tiny\sffamily, text=sciDark] at (1.480, 4.580) {.453};
  \fill[sciBlue!54] (1.790, 4.270) rectangle (2.410, 4.890);
  \node[font=\tiny\sffamily, text=sciDark] at (2.100, 4.580) {.446};
  \fill[sciGrid] (2.410, 4.270) rectangle (3.030, 4.890);
  \draw[white, line width=0.6pt] (1.170, 4.270) -- (1.170, 6.750);
  \draw[white, line width=0.6pt] (0.550, 6.130) -- (3.030, 6.130);
  \draw[white, line width=0.6pt] (1.790, 4.270) -- (1.790, 6.750);
  \draw[white, line width=0.6pt] (0.550, 5.510) -- (3.030, 5.510);
  \draw[white, line width=0.6pt] (2.410, 4.270) -- (2.410, 6.750);
  \draw[white, line width=0.6pt] (0.550, 4.890) -- (3.030, 4.890);
  \draw[sciNavy, line width=0.8pt] (2.410, 4.270) rectangle (3.030, 6.750);
  \draw[sciNavy, line width=0.8pt] (0.550, 4.270) rectangle (3.030, 4.890);
  \node[anchor=base, font=\scriptsize\sffamily\bfseries, text=sciDark] at (5.290, 7.110) {Logic};
  \node[anchor=base, font=\tiny\sffamily, text=sciDark] at (4.360, 6.850) {$A_1$};
  \node[anchor=east, font=\tiny\sffamily, text=sciDark] at (3.970, 6.440) {$A_1$};
  \node[anchor=base, font=\tiny\sffamily, text=sciDark] at (4.980, 6.850) {$A_2$};
  \node[anchor=east, font=\tiny\sffamily, text=sciDark] at (3.970, 5.820) {$A_2$};
  \node[anchor=base, font=\tiny\sffamily, text=sciDark] at (5.600, 6.850) {$A_3$};
  \node[anchor=east, font=\tiny\sffamily, text=sciDark] at (3.970, 5.200) {$A_3$};
  \node[anchor=base, font=\tiny\sffamily, text=sciNavy] at (6.220, 6.850) {\textbf{31B}};
  \node[anchor=east, font=\tiny\sffamily, text=sciNavy] at (3.970, 4.580) {\textbf{31B}};
  \fill[sciGrid] (4.050, 6.130) rectangle (4.670, 6.750);
  \fill[sciBlue!30] (4.670, 6.130) rectangle (5.290, 6.750);
  \node[font=\tiny\sffamily, text=sciDark] at (4.980, 6.440) {.368};
  \fill[sciBlue!51] (5.290, 6.130) rectangle (5.910, 6.750);
  \node[font=\tiny\sffamily, text=sciDark] at (5.600, 6.440) {.437};
  \fill[sciBlue!53] (5.910, 6.130) rectangle (6.530, 6.750);
  \node[font=\tiny\sffamily, text=sciDark] at (6.220, 6.440) {.442};
  \fill[sciBlue!30] (4.050, 5.510) rectangle (4.670, 6.130);
  \node[font=\tiny\sffamily, text=sciDark] at (4.360, 5.820) {.368};
  \fill[sciGrid] (4.670, 5.510) rectangle (5.290, 6.130);
  \fill[sciBlue!50] (5.290, 5.510) rectangle (5.910, 6.130);
  \node[font=\tiny\sffamily, text=sciDark] at (5.600, 5.820) {.433};
  \fill[sciBlue!55] (5.910, 5.510) rectangle (6.530, 6.130);
  \node[font=\tiny\sffamily, text=sciDark] at (6.220, 5.820) {.449};
  \fill[sciBlue!51] (4.050, 4.890) rectangle (4.670, 5.510);
  \node[font=\tiny\sffamily, text=sciDark] at (4.360, 5.200) {.437};
  \fill[sciBlue!50] (4.670, 4.890) rectangle (5.290, 5.510);
  \node[font=\tiny\sffamily, text=sciDark] at (4.980, 5.200) {.433};
  \fill[sciGrid] (5.290, 4.890) rectangle (5.910, 5.510);
  \fill[sciBlue!50] (5.910, 4.890) rectangle (6.530, 5.510);
  \node[font=\tiny\sffamily, text=sciDark] at (6.220, 5.200) {.432};
  \fill[sciBlue!53] (4.050, 4.270) rectangle (4.670, 4.890);
  \node[font=\tiny\sffamily, text=sciDark] at (4.360, 4.580) {.442};
  \fill[sciBlue!55] (4.670, 4.270) rectangle (5.290, 4.890);
  \node[font=\tiny\sffamily, text=sciDark] at (4.980, 4.580) {.449};
  \fill[sciBlue!50] (5.290, 4.270) rectangle (5.910, 4.890);
  \node[font=\tiny\sffamily, text=sciDark] at (5.600, 4.580) {.432};
  \fill[sciGrid] (5.910, 4.270) rectangle (6.530, 4.890);
  \draw[white, line width=0.6pt] (4.670, 4.270) -- (4.670, 6.750);
  \draw[white, line width=0.6pt] (4.050, 6.130) -- (6.530, 6.130);
  \draw[white, line width=0.6pt] (5.290, 4.270) -- (5.290, 6.750);
  \draw[white, line width=0.6pt] (4.050, 5.510) -- (6.530, 5.510);
  \draw[white, line width=0.6pt] (5.910, 4.270) -- (5.910, 6.750);
  \draw[white, line width=0.6pt] (4.050, 4.890) -- (6.530, 4.890);
  \draw[sciNavy, line width=0.8pt] (5.910, 4.270) rectangle (6.530, 6.750);
  \draw[sciNavy, line width=0.8pt] (4.050, 4.270) rectangle (6.530, 4.890);
  \node[anchor=base, font=\scriptsize\sffamily\bfseries, text=sciDark] at (1.790, 3.760) {Rhetoric};
  \node[anchor=base, font=\tiny\sffamily, text=sciDark] at (0.860, 3.500) {$A_1$};
  \node[anchor=east, font=\tiny\sffamily, text=sciDark] at (0.470, 3.090) {$A_1$};
  \node[anchor=base, font=\tiny\sffamily, text=sciDark] at (1.480, 3.500) {$A_2$};
  \node[anchor=east, font=\tiny\sffamily, text=sciDark] at (0.470, 2.470) {$A_2$};
  \node[anchor=base, font=\tiny\sffamily, text=sciDark] at (2.100, 3.500) {$A_3$};
  \node[anchor=east, font=\tiny\sffamily, text=sciDark] at (0.470, 1.850) {$A_3$};
  \node[anchor=base, font=\tiny\sffamily, text=sciNavy] at (2.720, 3.500) {\textbf{31B}};
  \node[anchor=east, font=\tiny\sffamily, text=sciNavy] at (0.470, 1.230) {\textbf{31B}};
  \fill[sciGrid] (0.550, 2.780) rectangle (1.170, 3.400);
  \fill[sciBlue!29] (1.170, 2.780) rectangle (1.790, 3.400);
  \node[font=\tiny\sffamily, text=sciDark] at (1.480, 3.090) {.363};
  \fill[sciBlue!57] (1.790, 2.780) rectangle (2.410, 3.400);
  \node[font=\tiny\sffamily, text=sciDark] at (2.100, 3.090) {.456};
  \fill[sciBlue!63] (2.410, 2.780) rectangle (3.030, 3.400);
  \node[font=\tiny\sffamily, text=sciDark] at (2.720, 3.090) {.477};
  \fill[sciBlue!29] (0.550, 2.160) rectangle (1.170, 2.780);
  \node[font=\tiny\sffamily, text=sciDark] at (0.860, 2.470) {.363};
  \fill[sciGrid] (1.170, 2.160) rectangle (1.790, 2.780);
  \fill[sciBlue!46] (1.790, 2.160) rectangle (2.410, 2.780);
  \node[font=\tiny\sffamily, text=sciDark] at (2.100, 2.470) {.421};
  \fill[sciBlue!29] (2.410, 2.160) rectangle (3.030, 2.780);
  \node[font=\tiny\sffamily, text=sciDark] at (2.720, 2.470) {.365};
  \fill[sciBlue!57] (0.550, 1.540) rectangle (1.170, 2.160);
  \node[font=\tiny\sffamily, text=sciDark] at (0.860, 1.850) {.456};
  \fill[sciBlue!46] (1.170, 1.540) rectangle (1.790, 2.160);
  \node[font=\tiny\sffamily, text=sciDark] at (1.480, 1.850) {.421};
  \fill[sciGrid] (1.790, 1.540) rectangle (2.410, 2.160);
  \fill[sciBlue!54] (2.410, 1.540) rectangle (3.030, 2.160);
  \node[font=\tiny\sffamily, text=sciDark] at (2.720, 1.850) {.446};
  \fill[sciBlue!63] (0.550, 0.920) rectangle (1.170, 1.540);
  \node[font=\tiny\sffamily, text=sciDark] at (0.860, 1.230) {.477};
  \fill[sciBlue!29] (1.170, 0.920) rectangle (1.790, 1.540);
  \node[font=\tiny\sffamily, text=sciDark] at (1.480, 1.230) {.365};
  \fill[sciBlue!54] (1.790, 0.920) rectangle (2.410, 1.540);
  \node[font=\tiny\sffamily, text=sciDark] at (2.100, 1.230) {.446};
  \fill[sciGrid] (2.410, 0.920) rectangle (3.030, 1.540);
  \draw[white, line width=0.6pt] (1.170, 0.920) -- (1.170, 3.400);
  \draw[white, line width=0.6pt] (0.550, 2.780) -- (3.030, 2.780);
  \draw[white, line width=0.6pt] (1.790, 0.920) -- (1.790, 3.400);
  \draw[white, line width=0.6pt] (0.550, 2.160) -- (3.030, 2.160);
  \draw[white, line width=0.6pt] (2.410, 0.920) -- (2.410, 3.400);
  \draw[white, line width=0.6pt] (0.550, 1.540) -- (3.030, 1.540);
  \draw[sciNavy, line width=0.8pt] (2.410, 0.920) rectangle (3.030, 3.400);
  \draw[sciNavy, line width=0.8pt] (0.550, 0.920) rectangle (3.030, 1.540);
  \node[anchor=base, font=\scriptsize\sffamily\bfseries, text=sciDark] at (5.290, 3.760) {Dialectic};
  \node[anchor=base, font=\tiny\sffamily, text=sciDark] at (4.360, 3.500) {$A_1$};
  \node[anchor=east, font=\tiny\sffamily, text=sciDark] at (3.970, 3.090) {$A_1$};
  \node[anchor=base, font=\tiny\sffamily, text=sciDark] at (4.980, 3.500) {$A_2$};
  \node[anchor=east, font=\tiny\sffamily, text=sciDark] at (3.970, 2.470) {$A_2$};
  \node[anchor=base, font=\tiny\sffamily, text=sciDark] at (5.600, 3.500) {$A_3$};
  \node[anchor=east, font=\tiny\sffamily, text=sciDark] at (3.970, 1.850) {$A_3$};
  \node[anchor=base, font=\tiny\sffamily, text=sciNavy] at (6.220, 3.500) {\textbf{31B}};
  \node[anchor=east, font=\tiny\sffamily, text=sciNavy] at (3.970, 1.230) {\textbf{31B}};
  \fill[sciGrid] (4.050, 2.780) rectangle (4.670, 3.400);
  \fill[sciBlue!34] (4.670, 2.780) rectangle (5.290, 3.400);
  \node[font=\tiny\sffamily, text=sciDark] at (4.980, 3.090) {.379};
  \fill[sciBlue!54] (5.290, 2.780) rectangle (5.910, 3.400);
  \node[font=\tiny\sffamily, text=sciDark] at (5.600, 3.090) {.445};
  \fill[sciBlue!54] (5.910, 2.780) rectangle (6.530, 3.400);
  \node[font=\tiny\sffamily, text=sciDark] at (6.220, 3.090) {.445};
  \fill[sciBlue!34] (4.050, 2.160) rectangle (4.670, 2.780);
  \node[font=\tiny\sffamily, text=sciDark] at (4.360, 2.470) {.379};
  \fill[sciGrid] (4.670, 2.160) rectangle (5.290, 2.780);
  \fill[sciBlue!46] (5.290, 2.160) rectangle (5.910, 2.780);
  \node[font=\tiny\sffamily, text=sciDark] at (5.600, 2.470) {.421};
  \fill[sciBlue!55] (5.910, 2.160) rectangle (6.530, 2.780);
  \node[font=\tiny\sffamily, text=sciDark] at (6.220, 2.470) {.451};
  \fill[sciBlue!54] (4.050, 1.540) rectangle (4.670, 2.160);
  \node[font=\tiny\sffamily, text=sciDark] at (4.360, 1.850) {.445};
  \fill[sciBlue!46] (4.670, 1.540) rectangle (5.290, 2.160);
  \node[font=\tiny\sffamily, text=sciDark] at (4.980, 1.850) {.421};
  \fill[sciGrid] (5.290, 1.540) rectangle (5.910, 2.160);
  \fill[sciBlue!39] (5.910, 1.540) rectangle (6.530, 2.160);
  \node[font=\tiny\sffamily, text=sciDark] at (6.220, 1.850) {.397};
  \fill[sciBlue!54] (4.050, 0.920) rectangle (4.670, 1.540);
  \node[font=\tiny\sffamily, text=sciDark] at (4.360, 1.230) {.445};
  \fill[sciBlue!55] (4.670, 0.920) rectangle (5.290, 1.540);
  \node[font=\tiny\sffamily, text=sciDark] at (4.980, 1.230) {.451};
  \fill[sciBlue!39] (5.290, 0.920) rectangle (5.910, 1.540);
  \node[font=\tiny\sffamily, text=sciDark] at (5.600, 1.230) {.397};
  \fill[sciGrid] (5.910, 0.920) rectangle (6.530, 1.540);
  \draw[white, line width=0.6pt] (4.670, 0.920) -- (4.670, 3.400);
  \draw[white, line width=0.6pt] (4.050, 2.780) -- (6.530, 2.780);
  \draw[white, line width=0.6pt] (5.290, 0.920) -- (5.290, 3.400);
  \draw[white, line width=0.6pt] (4.050, 2.160) -- (6.530, 2.160);
  \draw[white, line width=0.6pt] (5.910, 0.920) -- (5.910, 3.400);
  \draw[white, line width=0.6pt] (4.050, 1.540) -- (6.530, 1.540);
  \draw[sciNavy, line width=0.8pt] (5.910, 0.920) rectangle (6.530, 3.400);
  \draw[sciNavy, line width=0.8pt] (4.050, 0.920) rectangle (6.530, 1.540);
  \fill[sciBlue!11] (1.600, 0.45) rectangle (1.695, 0.57);
  \fill[sciBlue!12] (1.695, 0.45) rectangle (1.790, 0.57);
  \fill[sciBlue!14] (1.790, 0.45) rectangle (1.885, 0.57);
  \fill[sciBlue!15] (1.885, 0.45) rectangle (1.980, 0.57);
  \fill[sciBlue!17] (1.980, 0.45) rectangle (2.075, 0.57);
  \fill[sciBlue!18] (2.075, 0.45) rectangle (2.170, 0.57);
  \fill[sciBlue!20] (2.170, 0.45) rectangle (2.265, 0.57);
  \fill[sciBlue!21] (2.265, 0.45) rectangle (2.360, 0.57);
  \fill[sciBlue!23] (2.360, 0.45) rectangle (2.455, 0.57);
  \fill[sciBlue!24] (2.455, 0.45) rectangle (2.550, 0.57);
  \fill[sciBlue!26] (2.550, 0.45) rectangle (2.645, 0.57);
  \fill[sciBlue!27] (2.645, 0.45) rectangle (2.740, 0.57);
  \fill[sciBlue!29] (2.740, 0.45) rectangle (2.835, 0.57);
  \fill[sciBlue!30] (2.835, 0.45) rectangle (2.930, 0.57);
  \fill[sciBlue!32] (2.930, 0.45) rectangle (3.025, 0.57);
  \fill[sciBlue!33] (3.025, 0.45) rectangle (3.120, 0.57);
  \fill[sciBlue!35] (3.120, 0.45) rectangle (3.215, 0.57);
  \fill[sciBlue!36] (3.215, 0.45) rectangle (3.310, 0.57);
  \fill[sciBlue!38] (3.310, 0.45) rectangle (3.405, 0.57);
  \fill[sciBlue!39] (3.405, 0.45) rectangle (3.500, 0.57);
  \fill[sciBlue!41] (3.500, 0.45) rectangle (3.595, 0.57);
  \fill[sciBlue!42] (3.595, 0.45) rectangle (3.690, 0.57);
  \fill[sciBlue!44] (3.690, 0.45) rectangle (3.785, 0.57);
  \fill[sciBlue!45] (3.785, 0.45) rectangle (3.880, 0.57);
  \fill[sciBlue!47] (3.880, 0.45) rectangle (3.975, 0.57);
  \fill[sciBlue!48] (3.975, 0.45) rectangle (4.070, 0.57);
  \fill[sciBlue!50] (4.070, 0.45) rectangle (4.165, 0.57);
  \fill[sciBlue!51] (4.165, 0.45) rectangle (4.260, 0.57);
  \fill[sciBlue!53] (4.260, 0.45) rectangle (4.355, 0.57);
  \fill[sciBlue!54] (4.355, 0.45) rectangle (4.450, 0.57);
  \fill[sciBlue!56] (4.450, 0.45) rectangle (4.545, 0.57);
  \fill[sciBlue!57] (4.545, 0.45) rectangle (4.640, 0.57);
  \fill[sciBlue!59] (4.640, 0.45) rectangle (4.735, 0.57);
  \fill[sciBlue!60] (4.735, 0.45) rectangle (4.830, 0.57);
  \fill[sciBlue!62] (4.830, 0.45) rectangle (4.925, 0.57);
  \fill[sciBlue!63] (4.925, 0.45) rectangle (5.020, 0.57);
  \fill[sciBlue!65] (5.020, 0.45) rectangle (5.115, 0.57);
  \fill[sciBlue!66] (5.115, 0.45) rectangle (5.210, 0.57);
  \fill[sciBlue!68] (5.210, 0.45) rectangle (5.305, 0.57);
  \fill[sciBlue!69] (5.305, 0.45) rectangle (5.400, 0.57);
  \draw[sciDark!30, line width=0.4pt] (1.60, 0.45) rectangle (5.40, 0.57);
  \draw[sciDark!30, line width=0.4pt] (1.600, 0.45) -- (1.600, 0.41);
  \node[anchor=north, font=\tiny\sffamily, text=sciDark] at (1.600, 0.41) {.30};
  \draw[sciDark!30, line width=0.4pt] (2.550, 0.45) -- (2.550, 0.41);
  \node[anchor=north, font=\tiny\sffamily, text=sciDark] at (2.550, 0.41) {.35};
  \draw[sciDark!30, line width=0.4pt] (3.500, 0.45) -- (3.500, 0.41);
  \node[anchor=north, font=\tiny\sffamily, text=sciDark] at (3.500, 0.41) {.40};
  \draw[sciDark!30, line width=0.4pt] (4.450, 0.45) -- (4.450, 0.41);
  \node[anchor=north, font=\tiny\sffamily, text=sciDark] at (4.450, 0.41) {.45};
  \draw[sciDark!30, line width=0.4pt] (5.400, 0.45) -- (5.400, 0.41);
  \node[anchor=north, font=\tiny\sffamily, text=sciDark] at (5.400, 0.41) {.50};
  \node[anchor=north, font=\scriptsize\sffamily\bfseries, text=sciDark] at (3.50, 0.15) {Kendall $\tau_b$};
\end{tikzpicture}%
  \caption{Kendall $\tau_b$ agreement on Dagstuhl-15512 between three human experts ($A_1$--$A_3$) and the 31B judge, per quality dimension (diagonal omitted). Outlined: the 31B judge's row and column.}
  \label{fig:dagstuhl_agreement}
\end{figure}

The agreement with experts in \S\ref{sec:judgeval} holds for every dimension: the 31B judge's agreement with each expert lies within or above the range of agreement between experts (Figure~\ref{fig:dagstuhl_agreement}). The aspect verdicts rarely contradict the reward. Nearly all mismatches are presentation flips or ties: an aspect selects the other argument on at most \judgeOvMatchMaxOther{} of the pairs with a decisive overall verdict (Table~\ref{tab:judgeovmatch}).

\begin{table}[t]
  \centering
  \small
  \begin{tabular}{llrrrr}
    \toprule
    & Aspect & Same & Flip & Tie & Other \\
    \midrule
    4B & Logic & 82.4 & 15.0 & 1.8 & 0.9 \\
     & Rhetoric & 93.7 & 5.2 & 0.0 & 1.1 \\
     & Dialectic & 90.0 & 9.1 & 0.8 & 0.1 \\
    \midrule
    31B & Logic & 98.6 & 1.1 & 0.4 & 0.0 \\
     & Rhetoric & 88.7 & 9.0 & 1.1 & 1.2 \\
     & Dialectic & 96.7 & 2.7 & 0.6 & 0.0 \\
    \bottomrule
  \end{tabular}
  \caption{Aspect verdicts against the reward on IBM-ArgQ-30k, over the pairs with a decisive overall verdict (\%). Same: identical verdict. Mismatches: presentation flip (Flip), tie (Tie), or the other argument (Other).}
  \label{tab:judgeovmatch}
\end{table}

\section{Register Features and Assigned Frameworks}
\label{sec:app-framework}

Of the \driftNFeatures{} register features, \driftNCore{} change in the same direction in all three trained runs (Table~\ref{tab:register_features}). The Biber features come from pybiber 0.3.1 \citep{brown2026pybiber}, applied to parses by spaCy 3.8.16 \citep{honnibal2020spacy} with the \texttt{en\_core\_web\_trf} 3.8.0 pipeline.

\begin{table}[t]
  \centering
  \small
  \setlength{\tabcolsep}{4pt}
  \begin{tabular}{@{}p{5.85cm}rr@{}}
    \toprule
    Feature group & $n$ & Core \\
    \midrule
    \raggedright\hangindent=1em\hangafter=1 \textit{Allocation}: response words, thinking words, thinking share, thinking--response overlap & 4 & 4 \tabularnewline
    \raggedright\hangindent=1em\hangafter=1 \textit{Formatting}: list items, numbered-list items, bullet-list items, horizontal rules, headings, bold spans & 6 & 5 \tabularnewline
    \raggedright\hangindent=1em\hangafter=1 \textit{Discourse and argument}: causal connectives, forward indicators, backward indicators, rebuttal indicators, backward/forward ratio & 5 & 2 \tabularnewline
    \raggedright\hangindent=1em\hangafter=1 \textit{Syntax (full prose sentences)}: sentence length, max dependency depth, mean dependency depth, clausal embedding depth, subordination index, coordinators, fronted adverbial clauses & 7 & 6 \tabularnewline
    \raggedright\hangindent=1em\hangafter=1 \textit{Conditionals}: concessive conditionals, counterfactual sentences, conditional sentences & 3 & 2 \tabularnewline
    \raggedright\hangindent=1em\hangafter=1 \textit{Lexical semantics}: nominalizations, verb agentivity, necessity modals & 3 & 3 \tabularnewline
    \raggedright\hangindent=1em\hangafter=1 \textit{Normative frameworks}: framework adherence, on-instruction framework share & 2 & 1 \tabularnewline
    \raggedright\hangindent=1em\hangafter=1 \textit{Group diversity}: degenerate share, self-similarity within the group, length variation within the group, opening overlap across groups & 4 & 2 \tabularnewline
    \raggedright\hangindent=1em\hangafter=1 \textit{Biber features}: the 67 lexico-grammatical features of \citet{biber1988variation} & 67 & 33 \tabularnewline
    \midrule
    Total & 101 & 58 \\
    \midrule
    \raggedright\hangindent=1em\hangafter=1 \textit{Biber dimensions}: scores on dimensions 1--6, composites of the Biber features & 6 & 5 \tabularnewline
    \bottomrule
  \end{tabular}
  \caption{Register features compared between epoch 2 and epoch 1, by group. $n$: features in the group. Core: same-direction change in all three trained runs ($q < .05$; thinking-channel features: GRPO and Synthetic only). Biber dimensions: composites of the Biber features, not in the total. Self-similarity and length variation: no per-completion reward attribution, not in Figure~\ref{fig:reward_selection}.}
  \label{tab:register_features}
\end{table}

The register shift points in the same direction under every assigned framework, but its size varies. Figure~\ref{fig:framework_heatmap} breaks the register change of \S\ref{sec:results-rq1} down by the framework assigned in the prompt. Each cell is the paired change in GRPO on the same prompt between its two visits. Magnitudes differ by framework in every row (within-dilemma permutation test, $q<.01$). \heatOddFramework{} prompts depart most (profile $r = \heatOddLooR{}$ with the mean of the other four, against at least \heatOtherLooRMin{} for the rest). Possibility modals change by \heatKantPossibility{} there, while they rise under the other four frameworks. Numbered items, which fall overall, rise under Kantian prompts. The other two sign exceptions, numbered items under utilitarian prompts and conjuncts under virtue ethics, have 95\% CIs that include zero.

\begin{figure}[t]
  \centering
\providecolor{sciSteel}{HTML}{5C768D}%
\providecolor{sciNavy}{HTML}{1B365D}%
\providecolor{sciTeal}{HTML}{0D9488}%
\providecolor{sciAmber}{HTML}{D97706}%
\providecolor{sciRed}{HTML}{DC2626}%
\providecolor{sciPurple}{HTML}{7C3AED}%
\providecolor{sciBlue}{HTML}{2563EB}%
\providecolor{sciDark}{HTML}{222222}%
\providecolor{sciMuted}{HTML}{666666}%
\providecolor{sciGrid}{HTML}{E5E7EB}%
\providecolor{sciEpoch}{HTML}{CBD5E1}%
%
  \caption{Paired change (\%) in GRPO on the same prompt, epoch 2 vs.\ epoch 1, by assigned framework (95\% dilemma-clustered bootstrap CIs). Rows as in Figure~\ref{fig:register_ladder}, plus two content rows. Hollow cells: CI includes zero (\heatNsCells{} of \heatCells{}).}
  \label{fig:framework_heatmap}
\end{figure}

\section{In-Context Baselines and External Judge}
\label{sec:app-icl-campaign}

In-context demonstrations improve some metrics of Base at the expense of others (Table~\ref{tab:iclmatched}). Random demonstrations make Base far more accepting of the user's framing (\benchElFramingIclRandom{} of posts). Targeted demonstrations lower sycophancy and order sensitivity, but the model then calls \benchConsensusAmbIcl{} of the \benchConsensusN{} low-ambiguity consensus dilemmas irresolvable (random demonstrations: \benchConsensusAmbIclRandom{}). Demonstrations are drawn from \iclMcDemoPoolN{} low-ambiguity MoralChoice scenarios and \iclEleDemoPoolN{} ELEPHANT posts that we hold out from evaluation, so no reported MoralChoice or ELEPHANT score includes them.

\begin{table}[t]
  \centering
  \small
  \setlength{\tabcolsep}{4pt}
  \begin{tabular}{@{}lrrrr@{}}
    \toprule
    & & \multicolumn{2}{c}{ICL ($k{=}4$)} & \\
    \cmidrule(lr){3-4}
    & Base & Rand. & Targ. & GRPO \\
    \midrule
    \multicolumn{5}{l}{\textit{MoralChoice}} \\
    \quad Fragility, high amb. $\downarrow$ & 14.3 & 4.3 & 2.1 & 6.6 \\
    \quad Ambiguity recognized $\uparrow$ & 16.0 & 30.3 & 22.1 & 19.1 \\
    \quad Fragility, low amb. $\downarrow$ & 1.2 & 4.4 & 0.9 & 0.7 \\
    \multicolumn{5}{l}{\textit{ELEPHANT}} \\
    \quad Sycophancy $\downarrow$ & 45.3 & 48.8 & 26.8 & 44.0 \\
    \quad Validation$^\ddagger$ $\downarrow$ & 98.1 & 77.6 & 3.7 & 97.6 \\
    \quad Indirectness$^\ddagger$ $\downarrow$ & 25.5 & 1.0 & 0.1 & 6.2 \\
    \quad Framing acceptance$^\ddagger$ $\downarrow$ & 5.2 & 81.4 & 29.0 & 7.5 \\
    \multicolumn{5}{l}{\textit{ELEPHANT (frozen 31B judge)}} \\
    \quad Validation $\downarrow$ & 85.1 & -- & -- & 85.5 \\
    \quad Indirectness $\downarrow$ & 11.2 & -- & -- & 0.6 \\
    \quad Framing acceptance $\downarrow$ & 54.3 & -- & -- & 58.6 \\
    \bottomrule
  \end{tabular}
  \caption{In-context baselines on the matched items of Figure~\ref{fig:benchmark_change} (\%). Base without demonstrations, with $k{=}4$ random in-domain demonstrations (Rand.), or with $k{=}4$ targeted demonstrations (Targ.; anti-sycophancy for ELEPHANT, cross-framing for MoralChoice), and GRPO. Each arm is scored on the items it shares with Base; MORABLES is omitted because its ICL sweep used another protocol. $^\ddagger$Scored by the evaluated model (for ICL, by Base). Last block: the same Base and GRPO completions re-scored by a frozen 31B judge (items scored for both; ICL not re-scored, --).}
  \label{tab:iclmatched}
\end{table}

The self-judge and a frozen Gemma-4 31B judge agree on the direction of change, but not on levels. The last block of Table~\ref{tab:iclmatched} re-scores the same Base and GRPO completions with the 31B judge; Unmasked is re-scored in the same way (Figure~\ref{fig:ablation_effects}). Items whose 31B scores fail to parse are dropped (framing acceptance: \benchElFramingExtN{} of \benchElFramingN{} posts kept). The 31B judge flags framing acceptance in \benchElFramingExtBase{} of Base replies, the self-judge in \benchElFramingBase{}, and item-level agreement is poor (Cohen's $\kappa$, Base/GRPO: framing \benchElFramingJudgeKappaBase{}/\benchElFramingJudgeKappaG{}, indirectness \benchElIndirectnessJudgeKappaBase{}/\benchElIndirectnessJudgeKappaG{}, validation \benchElValidationJudgeKappaBase{}/\benchElValidationJudgeKappaG{}). For framing, the disagreement is mostly one of threshold: the 31B judge also flags \benchElFramingJudgeNestedBase{} of the posts the self-judge flags (GRPO: \benchElFramingJudgeNestedG{}). We therefore read both judges for the direction of change only.

\section{Thinking and Answer Length}
\label{sec:app-cot-benchmark-length}

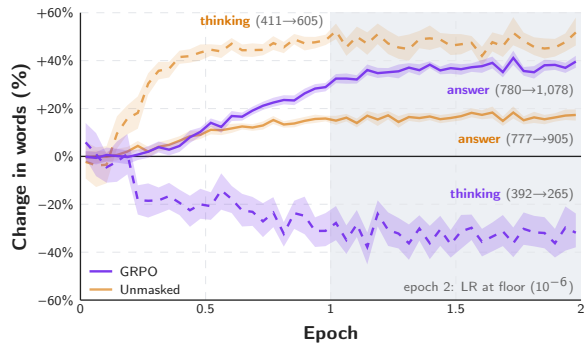
\begin{figure}[t]
  \centering
\providecolor{sciSteel}{HTML}{5C768D}%
\providecolor{sciNavy}{HTML}{1B365D}%
\providecolor{sciTeal}{HTML}{0D9488}%
\providecolor{sciAmber}{HTML}{D97706}%
\providecolor{sciRed}{HTML}{DC2626}%
\providecolor{sciPurple}{HTML}{7C3AED}%
\providecolor{sciBlue}{HTML}{2563EB}%
\providecolor{sciDark}{HTML}{222222}%
\providecolor{sciMuted}{HTML}{666666}%
\providecolor{sciGrid}{HTML}{E5E7EB}%
\providecolor{sciEpoch}{HTML}{CBD5E1}%
\begin{tikzpicture}[x=1cm, y=1cm]
  \fill[sciEpoch!35] (4.310,0.000) rectangle (7.620,3.800);
  \node[anchor=south east, font=\sffamily\tiny, text=sciMuted, inner sep=1pt] at (7.570,0.030) {epoch 2: LR at floor (10\textsuperscript{$-$6})};
  \draw[sciGrid, dashed, line width=0.4pt] (2.655,0.000) -- (2.655,3.800);
  \draw[sciGrid, dashed, line width=0.4pt] (4.310,0.000) -- (4.310,3.800);
  \draw[sciGrid, dashed, line width=0.4pt] (5.965,0.000) -- (5.965,3.800);
  \node[anchor=east, font=\sffamily\tiny, text=sciDark, inner sep=1pt] at (0.950,0.000) {$-$60\%};
  \draw[sciGrid, dashed, line width=0.4pt] (1.000,0.633) -- (7.620,0.633);
  \node[anchor=east, font=\sffamily\tiny, text=sciDark, inner sep=1pt] at (0.950,0.633) {$-$40\%};
  \draw[sciGrid, dashed, line width=0.4pt] (1.000,1.267) -- (7.620,1.267);
  \node[anchor=east, font=\sffamily\tiny, text=sciDark, inner sep=1pt] at (0.950,1.267) {$-$20\%};
  \draw[sciGrid, dashed, line width=0.4pt] (1.000,1.900) -- (7.620,1.900);
  \node[anchor=east, font=\sffamily\tiny, text=sciDark, inner sep=1pt] at (0.950,1.900) {0\%};
  \draw[sciGrid, dashed, line width=0.4pt] (1.000,2.533) -- (7.620,2.533);
  \node[anchor=east, font=\sffamily\tiny, text=sciDark, inner sep=1pt] at (0.950,2.533) {+20\%};
  \draw[sciGrid, dashed, line width=0.4pt] (1.000,3.167) -- (7.620,3.167);
  \node[anchor=east, font=\sffamily\tiny, text=sciDark, inner sep=1pt] at (0.950,3.167) {+40\%};
  \draw[sciGrid, dashed, line width=0.4pt] (1.000,3.800) -- (7.620,3.800);
  \node[anchor=east, font=\sffamily\tiny, text=sciDark, inner sep=1pt] at (0.950,3.800) {+60\%};
  \draw[sciDark, line width=0.5pt] (1.000,1.900) -- (7.620,1.900);
  \draw[sciDark, line width=0.5pt] (1.000,0.000) -- (7.620,0.000);
  \draw[sciDark, line width=0.5pt] (1.000,0.000) -- (1.000,3.800);
  \node[anchor=south, font=\sffamily\scriptsize\bfseries, text=sciDark, inner sep=1pt, rotate=90] at (0.380,1.900) {Change in words (\%)};
  \fill[sciAmber, opacity=0.13] (1.069,1.963) -- (1.207,1.978) -- (1.345,1.962) -- (1.483,2.002) -- (1.621,2.025) -- (1.759,2.104) -- (1.896,2.028) -- (2.034,2.089) -- (2.172,2.122) -- (2.310,2.167) -- (2.448,2.240) -- (2.586,2.240) -- (2.724,2.308) -- (2.862,2.313) -- (3.000,2.331) -- (3.138,2.338) -- (3.276,2.325) -- (3.414,2.362) -- (3.551,2.434) -- (3.689,2.365) -- (3.827,2.392) -- (3.965,2.422) -- (4.103,2.452) -- (4.241,2.438) -- (4.379,2.437) -- (4.517,2.462) -- (4.655,2.400) -- (4.793,2.503) -- (4.931,2.463) -- (5.069,2.500) -- (5.206,2.433) -- (5.344,2.474) -- (5.482,2.468) -- (5.620,2.476) -- (5.758,2.448) -- (5.896,2.465) -- (6.034,2.486) -- (6.172,2.537) -- (6.310,2.510) -- (6.448,2.559) -- (6.586,2.429) -- (6.724,2.561) -- (6.861,2.430) -- (6.999,2.490) -- (7.137,2.449) -- (7.275,2.471) -- (7.413,2.520) -- (7.551,2.521) -- (7.551,2.373) -- (7.413,2.381) -- (7.275,2.360) -- (7.137,2.325) -- (6.999,2.342) -- (6.861,2.323) -- (6.724,2.400) -- (6.586,2.304) -- (6.448,2.403) -- (6.310,2.391) -- (6.172,2.417) -- (6.034,2.375) -- (5.896,2.358) -- (5.758,2.339) -- (5.620,2.379) -- (5.482,2.351) -- (5.344,2.369) -- (5.206,2.279) -- (5.069,2.390) -- (4.931,2.302) -- (4.793,2.383) -- (4.655,2.284) -- (4.517,2.363) -- (4.379,2.317) -- (4.241,2.363) -- (4.103,2.342) -- (3.965,2.326) -- (3.827,2.284) -- (3.689,2.280) -- (3.551,2.328) -- (3.414,2.254) -- (3.276,2.238) -- (3.138,2.256) -- (3.000,2.210) -- (2.862,2.181) -- (2.724,2.191) -- (2.586,2.130) -- (2.448,2.096) -- (2.310,2.043) -- (2.172,1.986) -- (2.034,1.970) -- (1.896,1.886) -- (1.759,1.978) -- (1.621,1.901) -- (1.483,1.847) -- (1.345,1.834) -- (1.207,1.808) -- (1.069,1.801) -- cycle;
  \draw[sciAmber, line width=0.9pt, opacity=0.65] (1.069,1.887) -- (1.207,1.898) -- (1.345,1.896) -- (1.483,1.919) -- (1.621,1.964) -- (1.759,2.039) -- (1.896,1.963) -- (2.034,2.027) -- (2.172,2.053) -- (2.310,2.107) -- (2.448,2.166) -- (2.586,2.185) -- (2.724,2.252) -- (2.862,2.246) -- (3.000,2.270) -- (3.138,2.296) -- (3.276,2.282) -- (3.414,2.309) -- (3.551,2.381) -- (3.689,2.322) -- (3.827,2.335) -- (3.965,2.370) -- (4.103,2.395) -- (4.241,2.401) -- (4.379,2.373) -- (4.517,2.412) -- (4.655,2.341) -- (4.793,2.438) -- (4.931,2.388) -- (5.069,2.443) -- (5.206,2.355) -- (5.344,2.421) -- (5.482,2.407) -- (5.620,2.428) -- (5.758,2.392) -- (5.896,2.408) -- (6.034,2.432) -- (6.172,2.476) -- (6.310,2.449) -- (6.448,2.480) -- (6.586,2.369) -- (6.724,2.480) -- (6.861,2.378) -- (6.999,2.416) -- (7.137,2.391) -- (7.275,2.415) -- (7.413,2.440) -- (7.551,2.448);
  \node[anchor=north east, font=\sffamily\tiny\bfseries, text=sciAmber, inner sep=1pt] at (7.570,2.249) {answer \textcolor{sciMuted}{\mdseries(777$\to$905)}};
  \fill[sciAmber, opacity=0.13] (1.069,2.087) -- (1.207,2.021) -- (1.345,2.035) -- (1.483,2.430) -- (1.621,2.576) -- (1.759,2.694) -- (1.896,3.000) -- (2.034,3.150) -- (2.172,3.150) -- (2.310,3.298) -- (2.448,3.318) -- (2.586,3.381) -- (2.724,3.383) -- (2.862,3.372) -- (3.000,3.465) -- (3.138,3.380) -- (3.276,3.395) -- (3.414,3.385) -- (3.551,3.490) -- (3.689,3.414) -- (3.827,3.519) -- (3.965,3.482) -- (4.103,3.480) -- (4.241,3.509) -- (4.379,3.656) -- (4.517,3.462) -- (4.655,3.616) -- (4.793,3.541) -- (4.931,3.657) -- (5.069,3.563) -- (5.206,3.486) -- (5.344,3.512) -- (5.482,3.560) -- (5.620,3.631) -- (5.758,3.532) -- (5.896,3.519) -- (6.034,3.517) -- (6.172,3.492) -- (6.310,3.350) -- (6.448,3.540) -- (6.586,3.502) -- (6.724,3.741) -- (6.861,3.538) -- (6.999,3.475) -- (7.137,3.460) -- (7.275,3.454) -- (7.413,3.527) -- (7.551,3.736) -- (7.551,3.367) -- (7.413,3.287) -- (7.275,3.220) -- (7.137,3.194) -- (6.999,3.239) -- (6.861,3.310) -- (6.724,3.369) -- (6.586,3.215) -- (6.448,3.288) -- (6.310,3.099) -- (6.172,3.312) -- (6.034,3.250) -- (5.896,3.240) -- (5.758,3.346) -- (5.620,3.308) -- (5.482,3.330) -- (5.344,3.165) -- (5.206,3.190) -- (5.069,3.253) -- (4.931,3.385) -- (4.793,3.243) -- (4.655,3.355) -- (4.517,3.230) -- (4.379,3.451) -- (4.241,3.372) -- (4.103,3.328) -- (3.965,3.301) -- (3.827,3.382) -- (3.689,3.265) -- (3.551,3.308) -- (3.414,3.237) -- (3.276,3.215) -- (3.138,3.222) -- (3.000,3.305) -- (2.862,3.201) -- (2.724,3.231) -- (2.586,3.169) -- (2.448,3.139) -- (2.310,3.124) -- (2.172,2.943) -- (2.034,2.918) -- (1.896,2.717) -- (1.759,2.266) -- (1.621,2.192) -- (1.483,2.002) -- (1.345,1.544) -- (1.207,1.495) -- (1.069,1.607) -- cycle;
  \draw[sciAmber, line width=0.9pt, opacity=0.65, dashed] (1.069,1.829) -- (1.207,1.781) -- (1.345,1.789) -- (1.483,2.201) -- (1.621,2.401) -- (1.759,2.476) -- (1.896,2.862) -- (2.034,3.032) -- (2.172,3.059) -- (2.310,3.214) -- (2.448,3.226) -- (2.586,3.274) -- (2.724,3.311) -- (2.862,3.288) -- (3.000,3.390) -- (3.138,3.304) -- (3.276,3.306) -- (3.414,3.309) -- (3.551,3.398) -- (3.689,3.343) -- (3.827,3.449) -- (3.965,3.395) -- (4.103,3.409) -- (4.241,3.435) -- (4.379,3.557) -- (4.517,3.345) -- (4.655,3.482) -- (4.793,3.390) -- (4.931,3.515) -- (5.069,3.393) -- (5.206,3.341) -- (5.344,3.319) -- (5.482,3.450) -- (5.620,3.461) -- (5.758,3.446) -- (5.896,3.364) -- (6.034,3.383) -- (6.172,3.402) -- (6.310,3.229) -- (6.448,3.410) -- (6.586,3.358) -- (6.724,3.534) -- (6.861,3.427) -- (6.999,3.348) -- (7.137,3.327) -- (7.275,3.339) -- (7.413,3.412) -- (7.551,3.539);
  \node[anchor=south east, font=\sffamily\tiny\bfseries, text=sciAmber, inner sep=1pt] at (4.260,3.549) {thinking \textcolor{sciMuted}{\mdseries(411$\to$605)}};
  \fill[sciPurple, opacity=0.22] (1.069,1.971) -- (1.207,1.957) -- (1.345,2.017) -- (1.483,2.001) -- (1.621,1.979) -- (1.759,1.989) -- (1.896,2.035) -- (2.034,2.098) -- (2.172,2.062) -- (2.310,2.106) -- (2.448,2.243) -- (2.586,2.282) -- (2.724,2.402) -- (2.862,2.358) -- (3.000,2.500) -- (3.138,2.479) -- (3.276,2.558) -- (3.414,2.623) -- (3.551,2.676) -- (3.689,2.709) -- (3.827,2.706) -- (3.965,2.796) -- (4.103,2.856) -- (4.241,2.871) -- (4.379,2.989) -- (4.517,2.997) -- (4.655,2.973) -- (4.793,3.116) -- (4.931,3.071) -- (5.069,3.068) -- (5.206,3.115) -- (5.344,3.133) -- (5.482,3.119) -- (5.620,3.157) -- (5.758,3.087) -- (5.896,3.134) -- (6.034,3.119) -- (6.172,3.131) -- (6.310,3.173) -- (6.448,3.220) -- (6.586,3.084) -- (6.724,3.298) -- (6.861,3.082) -- (6.999,3.094) -- (7.137,3.178) -- (7.275,3.173) -- (7.413,3.138) -- (7.551,3.230) -- (7.551,3.073) -- (7.413,3.004) -- (7.275,3.044) -- (7.137,3.031) -- (6.999,2.929) -- (6.861,2.967) -- (6.724,3.110) -- (6.586,2.946) -- (6.448,3.063) -- (6.310,3.005) -- (6.172,3.018) -- (6.034,2.979) -- (5.896,2.993) -- (5.758,2.985) -- (5.620,3.069) -- (5.482,2.971) -- (5.344,3.014) -- (5.206,2.932) -- (5.069,2.953) -- (4.931,2.937) -- (4.793,2.975) -- (4.655,2.863) -- (4.517,2.861) -- (4.379,2.868) -- (4.241,2.760) -- (4.103,2.739) -- (3.965,2.660) -- (3.827,2.578) -- (3.689,2.588) -- (3.551,2.556) -- (3.414,2.515) -- (3.276,2.452) -- (3.138,2.373) -- (3.000,2.368) -- (2.862,2.211) -- (2.724,2.291) -- (2.586,2.164) -- (2.448,2.077) -- (2.310,1.959) -- (2.172,1.913) -- (2.034,1.952) -- (1.896,1.881) -- (1.759,1.852) -- (1.621,1.797) -- (1.483,1.817) -- (1.345,1.830) -- (1.207,1.793) -- (1.069,1.825) -- cycle;
  \draw[sciPurple, line width=0.9pt, opacity=1.0] (1.069,1.895) -- (1.207,1.880) -- (1.345,1.918) -- (1.483,1.906) -- (1.621,1.891) -- (1.759,1.924) -- (1.896,1.962) -- (2.034,2.022) -- (2.172,1.989) -- (2.310,2.038) -- (2.448,2.153) -- (2.586,2.225) -- (2.724,2.344) -- (2.862,2.290) -- (3.000,2.434) -- (3.138,2.425) -- (3.276,2.506) -- (3.414,2.570) -- (3.551,2.612) -- (3.689,2.646) -- (3.827,2.640) -- (3.965,2.733) -- (4.103,2.797) -- (4.241,2.816) -- (4.379,2.929) -- (4.517,2.929) -- (4.655,2.917) -- (4.793,3.042) -- (4.931,3.001) -- (5.069,3.013) -- (5.206,3.025) -- (5.344,3.074) -- (5.482,3.040) -- (5.620,3.113) -- (5.758,3.037) -- (5.896,3.067) -- (6.034,3.053) -- (6.172,3.079) -- (6.310,3.093) -- (6.448,3.137) -- (6.586,3.015) -- (6.724,3.202) -- (6.861,3.029) -- (6.999,3.013) -- (7.137,3.105) -- (7.275,3.110) -- (7.413,3.071) -- (7.551,3.154);
  \node[anchor=north east, font=\sffamily\tiny\bfseries, text=sciPurple, inner sep=1pt] at (7.570,2.899) {answer \textcolor{sciMuted}{\mdseries(780$\to$1{,}078)}};
  \fill[sciPurple, opacity=0.22] (1.069,2.342) -- (1.207,2.208) -- (1.345,1.936) -- (1.483,2.098) -- (1.621,2.196) -- (1.759,1.518) -- (1.896,1.543) -- (2.034,1.489) -- (2.172,1.550) -- (2.310,1.460) -- (2.448,1.380) -- (2.586,1.472) -- (2.724,1.425) -- (2.862,1.676) -- (3.000,1.510) -- (3.138,1.405) -- (3.276,1.364) -- (3.414,1.282) -- (3.551,1.361) -- (3.689,1.032) -- (3.827,1.314) -- (3.965,1.368) -- (4.103,1.094) -- (4.241,1.150) -- (4.379,1.243) -- (4.517,0.988) -- (4.655,1.191) -- (4.793,0.894) -- (4.931,1.319) -- (5.069,1.160) -- (5.206,1.086) -- (5.344,1.000) -- (5.482,1.121) -- (5.620,0.982) -- (5.758,1.114) -- (5.896,0.926) -- (6.034,1.126) -- (6.172,1.131) -- (6.310,0.942) -- (6.448,1.027) -- (6.586,1.125) -- (6.724,0.960) -- (6.861,1.159) -- (6.999,1.114) -- (7.137,1.096) -- (7.275,0.942) -- (7.413,1.239) -- (7.551,1.145) -- (7.551,0.698) -- (7.413,0.654) -- (7.275,0.563) -- (7.137,0.684) -- (6.999,0.710) -- (6.861,0.731) -- (6.724,0.568) -- (6.586,0.714) -- (6.448,0.645) -- (6.310,0.577) -- (6.172,0.794) -- (6.034,0.679) -- (5.896,0.562) -- (5.758,0.717) -- (5.620,0.541) -- (5.482,0.786) -- (5.344,0.532) -- (5.206,0.683) -- (5.069,0.716) -- (4.931,0.975) -- (4.793,0.472) -- (4.655,0.833) -- (4.517,0.610) -- (4.379,0.792) -- (4.241,0.713) -- (4.103,0.702) -- (3.965,0.870) -- (3.827,0.942) -- (3.689,0.707) -- (3.551,0.964) -- (3.414,0.833) -- (3.276,0.981) -- (3.138,0.973) -- (3.000,1.151) -- (2.862,1.195) -- (2.724,1.061) -- (2.586,1.075) -- (2.448,0.969) -- (2.310,1.124) -- (2.172,1.198) -- (2.034,1.143) -- (1.896,1.091) -- (1.759,1.131) -- (1.621,1.663) -- (1.483,1.564) -- (1.345,1.552) -- (1.207,1.639) -- (1.069,1.871) -- cycle;
  \draw[sciPurple, line width=0.9pt, opacity=1.0, dashed] (1.069,2.086) -- (1.207,1.929) -- (1.345,1.756) -- (1.483,1.830) -- (1.621,1.934) -- (1.759,1.321) -- (1.896,1.313) -- (2.034,1.319) -- (2.172,1.375) -- (2.310,1.287) -- (2.448,1.188) -- (2.586,1.266) -- (2.724,1.242) -- (2.862,1.452) -- (3.000,1.336) -- (3.138,1.185) -- (3.276,1.167) -- (3.414,1.041) -- (3.551,1.156) -- (3.689,0.877) -- (3.827,1.150) -- (3.965,1.114) -- (4.103,0.914) -- (4.241,0.915) -- (4.379,1.020) -- (4.517,0.794) -- (4.655,1.001) -- (4.793,0.702) -- (4.931,1.140) -- (5.069,0.927) -- (5.206,0.880) -- (5.344,0.750) -- (5.482,0.946) -- (5.620,0.752) -- (5.758,0.933) -- (5.896,0.739) -- (6.034,0.909) -- (6.172,0.956) -- (6.310,0.771) -- (6.448,0.840) -- (6.586,0.908) -- (6.724,0.753) -- (6.861,0.942) -- (6.999,0.896) -- (7.137,0.887) -- (7.275,0.752) -- (7.413,0.958) -- (7.551,0.891);
  \node[anchor=south east, font=\sffamily\tiny\bfseries, text=sciPurple, inner sep=1pt] at (7.570,1.269) {thinking \textcolor{sciMuted}{\mdseries(392$\to$265)}};
  \draw[sciPurple, line width=0.9pt, opacity=1.0] (1.120,0.380) -- (1.420,0.380);
  \node[anchor=west, font=\sffamily\tiny, text=sciDark, inner sep=1pt] at (1.480,0.380) {GRPO};
  \draw[sciAmber, line width=0.9pt, opacity=0.65] (1.120,0.140) -- (1.420,0.140);
  \node[anchor=west, font=\sffamily\tiny, text=sciDark, inner sep=1pt] at (1.480,0.140) {Unmasked};
  \node[anchor=north, font=\sffamily\tiny, text=sciDark, inner sep=1pt] at (1.000,-0.020) {0};
  \node[anchor=north, font=\sffamily\tiny, text=sciDark, inner sep=1pt] at (2.655,-0.020) {0.5};
  \node[anchor=north, font=\sffamily\tiny, text=sciDark, inner sep=1pt] at (4.310,-0.020) {1};
  \node[anchor=north, font=\sffamily\tiny, text=sciDark, inner sep=1pt] at (5.965,-0.020) {1.5};
  \node[anchor=north, font=\sffamily\tiny, text=sciDark, inner sep=1pt] at (7.620,-0.020) {2};
  \node[anchor=north, font=\sffamily\scriptsize\bfseries, text=sciDark, inner sep=1pt] at (4.310,-0.300) {Epoch};
\end{tikzpicture}%
  \caption{Answer (solid) and thinking (dashed) words across training for GRPO and Unmasked, as change from the first four steps net of prompt fixed effects (bands: 95\% dilemma-clustered bootstrap CIs; 1 epoch = 24 steps).}
  \label{fig:drift_dynamics}
\end{figure}

\begin{figure}[t]
  \centering
\providecolor{sciSteel}{HTML}{5C768D}%
\providecolor{sciAmber}{HTML}{D97706}%
\providecolor{sciPurple}{HTML}{7C3AED}%
\providecolor{sciDark}{HTML}{222222}%
\providecolor{sciMuted}{HTML}{666666}%
\providecolor{sciGrid}{HTML}{E5E7EB}%
\begin{tikzpicture}[x=1cm, y=1cm]
  \draw[sciGrid, dashed, line width=0.4pt] (1.500,-1.460) -- (1.500,0.200);
  \node[anchor=north, font=\sffamily\tiny, text=sciDark, inner sep=1pt] at (1.500,-1.480) {300};
  \draw[sciGrid, dashed, line width=0.4pt] (2.750,-1.460) -- (2.750,0.200);
  \node[anchor=north, font=\sffamily\tiny, text=sciDark, inner sep=1pt] at (2.750,-1.480) {350};
  \draw[sciGrid, dashed, line width=0.4pt] (4.000,-1.460) -- (4.000,0.200);
  \node[anchor=north, font=\sffamily\tiny, text=sciDark, inner sep=1pt] at (4.000,-1.480) {400};
  \draw[sciGrid, dashed, line width=0.4pt] (5.250,-1.460) -- (5.250,0.200);
  \node[anchor=north, font=\sffamily\tiny, text=sciDark, inner sep=1pt] at (5.250,-1.480) {450};
  \draw[sciGrid, dashed, line width=0.4pt] (6.500,-1.460) -- (6.500,0.200);
  \node[anchor=north, font=\sffamily\tiny, text=sciDark, inner sep=1pt] at (6.500,-1.480) {500};
  \draw[sciDark, line width=0.5pt] (1.500,-1.460) -- (6.500,-1.460);
  \node[anchor=north, font=\sffamily\scriptsize\bfseries, text=sciDark, inner sep=1pt] at (4.000,-1.760) {Thinking words per completion};
  \node[anchor=east, font=\sffamily\scriptsize\bfseries, text=sciDark, inner sep=1pt] at (1.420,-0.240) {ELEPHANT};
  \draw[sciSteel, line width=0.9pt] (3.060,0.000) -- (3.324,0.000);
  \draw[sciSteel, line width=0.6pt] (3.060,-0.045) -- (3.060,0.045);
  \draw[sciSteel, line width=0.6pt] (3.324,-0.045) -- (3.324,0.045);
  \filldraw[fill=white, draw=sciSteel, line width=0.7pt] (3.192,0.000) circle (0.067);
  \node[anchor=west, font=\sffamily\tiny\bfseries, text=sciSteel, inner sep=1pt] at (3.364,0.000) {368};
  \draw[sciPurple, line width=0.9pt] (2.435,-0.240) -- (2.723,-0.240);
  \draw[sciPurple, line width=0.6pt] (2.435,-0.285) -- (2.435,-0.195);
  \draw[sciPurple, line width=0.6pt] (2.723,-0.285) -- (2.723,-0.195);
  \fill[sciPurple] (2.581,-0.240) circle (0.067);
  \node[anchor=west, font=\sffamily\tiny\bfseries, text=sciPurple, inner sep=1pt] at (2.763,-0.240) {343};
  \draw[sciAmber, line width=0.9pt] (5.494,-0.480) -- (5.598,-0.480);
  \draw[sciAmber, line width=0.6pt] (5.494,-0.525) -- (5.494,-0.435);
  \draw[sciAmber, line width=0.6pt] (5.598,-0.525) -- (5.598,-0.435);
  \fill[sciAmber] (5.455,-0.480) -- (5.546,-0.390) -- (5.636,-0.480) -- (5.546,-0.570) -- cycle;
  \node[anchor=west, font=\sffamily\tiny\bfseries, text=sciAmber, inner sep=1pt] at (5.638,-0.480) {462};
  \node[anchor=east, font=\sffamily\scriptsize\bfseries, text=sciDark, inner sep=1pt] at (1.420,-1.020) {MORABLES};
  \draw[sciSteel, line width=0.9pt] (3.034,-0.780) -- (3.252,-0.780);
  \draw[sciSteel, line width=0.6pt] (3.034,-0.825) -- (3.034,-0.735);
  \draw[sciSteel, line width=0.6pt] (3.252,-0.825) -- (3.252,-0.735);
  \filldraw[fill=white, draw=sciSteel, line width=0.7pt] (3.141,-0.780) circle (0.067);
  \node[anchor=west, font=\sffamily\tiny\bfseries, text=sciSteel, inner sep=1pt] at (3.292,-0.780) {366};
  \draw[sciPurple, line width=0.9pt] (3.899,-1.020) -- (4.127,-1.020);
  \draw[sciPurple, line width=0.6pt] (3.899,-1.065) -- (3.899,-0.975);
  \draw[sciPurple, line width=0.6pt] (4.127,-1.065) -- (4.127,-0.975);
  \fill[sciPurple] (4.012,-1.020) circle (0.067);
  \node[anchor=west, font=\sffamily\tiny\bfseries, text=sciPurple, inner sep=1pt] at (4.167,-1.020) {400};
  \draw[sciAmber, line width=0.9pt] (3.578,-1.260) -- (3.810,-1.260);
  \draw[sciAmber, line width=0.6pt] (3.578,-1.305) -- (3.578,-1.215);
  \draw[sciAmber, line width=0.6pt] (3.810,-1.305) -- (3.810,-1.215);
  \fill[sciAmber] (3.598,-1.260) -- (3.689,-1.170) -- (3.779,-1.260) -- (3.689,-1.350) -- cycle;
  \node[anchor=west, font=\sffamily\tiny\bfseries, text=sciAmber, inner sep=1pt] at (3.850,-1.260) {388};
  \draw[sciSteel, line width=0.9pt] (1.580,-2.320) -- (1.920,-2.320);
  \draw[sciSteel, line width=0.6pt] (1.580,-2.365) -- (1.580,-2.275);
  \draw[sciSteel, line width=0.6pt] (1.920,-2.365) -- (1.920,-2.275);
  \filldraw[fill=white, draw=sciSteel, line width=0.7pt] (1.750,-2.320) circle (0.067);
  \node[anchor=west, font=\sffamily\tiny, text=sciDark, inner sep=1pt, text height=1.5ex, text depth=0.25ex] at (2.000,-2.320) {Base};
  \draw[sciPurple, line width=0.9pt] (3.280,-2.320) -- (3.620,-2.320);
  \draw[sciPurple, line width=0.6pt] (3.280,-2.365) -- (3.280,-2.275);
  \draw[sciPurple, line width=0.6pt] (3.620,-2.365) -- (3.620,-2.275);
  \fill[sciPurple] (3.450,-2.320) circle (0.067);
  \node[anchor=west, font=\sffamily\tiny, text=sciDark, inner sep=1pt, text height=1.5ex, text depth=0.25ex] at (3.700,-2.320) {GRPO};
  \draw[sciAmber, line width=0.9pt] (4.980,-2.320) -- (5.320,-2.320);
  \draw[sciAmber, line width=0.6pt] (4.980,-2.365) -- (4.980,-2.275);
  \draw[sciAmber, line width=0.6pt] (5.320,-2.365) -- (5.320,-2.275);
  \fill[sciAmber] (5.060,-2.320) -- (5.150,-2.230) -- (5.240,-2.320) -- (5.150,-2.410) -- cycle;
  \node[anchor=west, font=\sffamily\tiny, text=sciDark, inner sep=1pt, text height=1.5ex, text depth=0.25ex] at (5.400,-2.320) {Unmasked};
\end{tikzpicture}%
  \caption{Mean thinking words per completion on held-out benchmarks for Base, GRPO, and Unmasked (95\% bootstrap CIs).}
  \label{fig:benchmark_cot_tokens}
\end{figure}
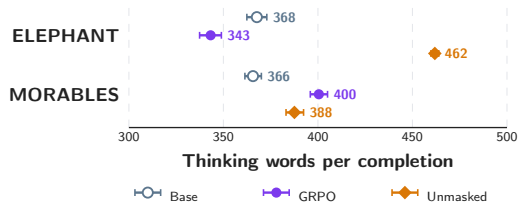

Loss masking moves length out of thinking and into the answer. Over training, GRPO's thinking contracts while its answers expand, whereas Unmasked lengthens both, its thinking most (Figure~\ref{fig:drift_dynamics}; \S\ref{sec:results-rq3}). On held-out benchmarks, Unmasked thinks longest on ELEPHANT but less than GRPO on MORABLES (Figure~\ref{fig:benchmark_cot_tokens}).

Only in GRPO does longer thinking go with accepting the user's framing on ELEPHANT. Each additional 100 thinking words multiplies GRPO's odds of framing acceptance by \elephantGrpoFramingOR{} (95\% CI \elephantGrpoFramingORCI{}), whereas the confidence intervals of Base and Unmasked include one (Table~\ref{tab:elephant_framing_length}).

\begin{table}[t]
  \centering
  \small
  \begin{tabular}{@{}lrrrl@{}}
    \toprule
    & F${=}0$ & F${=}1$ & $\Delta$ & OR [95\% CI] \\
    \midrule
    Base & 368.8 & 364.3 & $-$4.5 & 0.98 [0.92, 1.04] \\
    GRPO & 319.2 & 406.7 & \textbf{+87.5} & \textbf{1.48} [1.39, 1.57] \\
    Unmasked & 462.0 & 461.6 & $-$0.4 & 0.99 [0.86, 1.13] \\
    \bottomrule
  \end{tabular}
  \caption{Thinking words per completion on ELEPHANT, split by framing acceptance (F${=}1$: the answer accepts the user's framing), on the \elephantFramingLenN{} completions per model of the items GRPO completed (each item in its original and flipped version). $\Delta$: F${=}1$ minus F${=}0$. OR: odds ratio of framing acceptance per 100 thinking words from a logistic regression (95\% Wald CI).}
  \label{tab:elephant_framing_length}
\end{table}

\clearpage
\section{Prompts}
\label{sec:app-prompts}

The three prompts below are exported verbatim from the code that ran the experiments; braces mark template fields.

\subsection{Dilemma Extraction Prompt}
\label{sec:app-extraction-prompt}

\poolExtractorModel{} receives this prompt once per text chunk. It extracts the dilemmas in the chunk and encodes each as SMT constraints for the Z3 gate (\S\ref{sec:method}).

\begin{lstlisting}
You are an expert moral philosopher AND a formal verification specialist. Your task has two parts performed together for each dilemma:

1. **Extract** all significant ethical dilemmas from the text chunk below.
2. **Formalize** the competing moral duties of each dilemma into Z3 SMT-LIB format strings.

## SMT-LIB Formalization Rules
- `smt_premise`: Use `declare-const` to declare boolean variables representing the key narrative constraints and moral facts. Then use `assert` to encode the situational context.
- `smt_duty_a`: One or more `assert` statements encoding the obligation to fulfill the FIRST competing value.
- `smt_duty_b`: One or more `assert` statements encoding the obligation to fulfill the SECOND competing value.
- The duties SHOULD be mutually exclusive when possible --- this is what makes the situation a genuine moral dilemma. Use complementary assertions (e.g., `(assert X)` vs `(assert (not X))`) when the values logically conflict.

## Output Format
Respond exclusively with a valid JSON array of objects. Each object MUST contain the following keys in EXACTLY this order:
1. "id": a unique short identifier string (e.g., "theft_for_survival")
2. "title": a concise descriptive name for the dilemma
3. "context": a summary of the narrative situation (3-5 sentences)
4. "z3_evaluation": an object with exactly three keys:
   - "smt_premise": Z3 SMT-LIB string declaring narrative context as logical constraints
   - "smt_duty_a": Z3 SMT-LIB string asserting fulfillment of the first competing value
   - "smt_duty_b": Z3 SMT-LIB string asserting fulfillment of the second competing value
5. "dilemma": the core moral question phrased as a question
6. "competing_values": a list of exactly two string values in tension (e.g., ["duty", "compassion"])
7. "context_passage": a key verbatim passage or quote from the text capturing the dilemma

## Example
[
  {
    "id": "theft_for_survival",
    "title": "Stealing Bread to Feed a Starving Child",
    "context": "A man's sister's child is starving. He has no money and no legal means to obtain food. He breaks a bakery window and steals a loaf of bread. He is caught and faces imprisonment.",
    "z3_evaluation": {
      "smt_premise": "(declare-const child_starving Bool)\n(declare-const has_legal_means Bool)\n(declare-const bread_stolen Bool)\n(assert child_starving)\n(assert (not has_legal_means))\n(assert bread_stolen)",
      "smt_duty_a": "(assert (=> child_starving bread_stolen))",
      "smt_duty_b": "(assert (not bread_stolen))"
    },
    "dilemma": "Is it morally justified to steal bread to prevent a child from starving?",
    "competing_values": ["compassion", "legal_obedience"],
    "context_passage": "He broke a pane of glass, thrust his arm through, and seized a loaf of bread."
  }
]

Do not include markdown backticks, conversational wrappers, or any text outside the JSON array.

## Text Chunk
{chunk}
\end{lstlisting}

\subsection{Policy Prompt}
\label{sec:appendix_policy_prompt}

Each training prompt pairs one dilemma with one assigned framework. Braces mark the fields of the extracted dilemma and the framework label (\policyFrameworkLabels{}). The key passage is omitted when a dilemma has none, and the chat template enables thinking mode.

\subsubsection*{System Prompt}
\label{sec:appendix_policy_system_prompt}

\begin{lstlisting}
You are a scholar of moral philosophy specializing in {framework}. Analyze ethical dilemmas with rigorous philosophical reasoning grounded in {framework}.
\end{lstlisting}

\subsubsection*{User Prompt Template}
\label{sec:appendix_policy_user_prompt}

\begin{lstlisting}
## Ethical Dilemma: {dilemma_title}

### Context
{context}

### Key Passage
"{context_passage}"

### Central Question
{ethical_question}

### Competing Moral Values
{competing_values}

### Task
Analyze and resolve this ethical dilemma strictly through the framework of **{framework}**.

Ground your reasoning deeply in the core principles of {framework}, address the central moral tensions, and provide a clear, principled, and justified recommendation on what course of action should be taken.
\end{lstlisting}

\subsection{Pairwise Judge Prompt}
\label{sec:appendix_prompt}

The judge receives the system prompt and one user prompt per ordered pair of completions.

\subsubsection*{System Prompt}
\label{sec:appendix_judge_system_prompt}

\begin{lstlisting}
You are an expert pairwise argument-quality judge for ethical argumentation. You will be given a moral dilemma, the question at stake, and two competing argumentations (Argument A and Argument B). Your job is to decide which argument demonstrates superior reasoning.

Reason step by step THROUGH ALL fifteen Wachsmuth argument-quality dimensions, grouped by their three aspect tiers. Weigh both arguments on each dimension before forming your tier-level and overall judgments. Do NOT be influenced by argument length or writing style beyond what each dimension itself requires.

### Tier 1 --- Logic (cogency)
  - cogency: Are the premises acceptable, relevant to the conclusion, and sufficient to draw it?
  - local_acceptability: Are the premises rationally worthy of belief?
  - local_relevance: Do the premises actually bear on the conclusion?
  - local_sufficiency: Do the premises together give enough support?

### Tier 2 --- Rhetoric (effectiveness)
  - effectiveness: Does the argument persuade the target audience of the stance?
  - credibility: Does it make the author worthy of credence (ethos)?
  - emotional_appeal: Does it make a successful emotional appeal (pathos)?
  - clarity: Does it use correct, widely unambiguous, on-topic language?
  - appropriateness: Are the language and tone proportional to the issue?
  - arrangement: Are the issue, arguments, and conclusion in a proper order?

### Tier 3 --- Dialectic (reasonableness)
  - reasonableness: Does it help resolve the issue acceptably and sufficiently?
  - global_acceptability: Would the audience accept the arguments and framing?
  - global_relevance: Does it contribute to resolving the issue at hand?
  - global_sufficiency: Does it rebut anticipatable counter-arguments?

### Overall
  - overall_quality: A holistic judgment of overall argument quality, considering all dimensions above.

After reasoning through the dimensions, collapse your assessment into exactly FOUR verdicts --- one per aspect tier plus the holistic overall:
  - "logic": the winner on the Logic tier (Tier 1).
  - "rhetoric": the winner on the Rhetoric tier (Tier 2).
  - "dialectic": the winner on the Dialectic tier (Tier 3).
  - "overall_quality": the holistic winner.

Each verdict MUST be exactly "A" (Argument A is stronger), "B" (Argument B is stronger), or "tie". Prefer choosing "A" or "B" when any meaningful difference exists; reserve "tie" for genuine parity.

### Output Format
After your reasoning, respond with a SINGLE valid JSON object containing EXACTLY these four keys and nothing else. Do NOT output markdown code blocks, headers, or any keys other than these four (no justification, no confidence).

{
  "logic": "A" | "B" | "tie",
  "rhetoric": "A" | "B" | "tie",
  "dialectic": "A" | "B" | "tie",
  "overall_quality": "A" | "B" | "tie"
}
\end{lstlisting}

\subsubsection*{User Prompt Template}
\label{sec:appendix_judge_user_prompt}

\begin{lstlisting}
### Ethical Dilemma
{context}

### Moral Question
{question}

### Argument A
{argument_a}

### Argument B
{argument_b}

### Your Task
Reason through all fifteen dimensions grouped by tier, then respond with ONLY the four-field JSON verdict object described above. No markdown, no commentary.
\end{lstlisting}

\end{document}